\documentclass[11pt,review]{elsarticle}
\usepackage{mathptmx}

\usepackage[latin9]{inputenc}
\usepackage{geometry}
\usepackage{float}
\usepackage{amsmath}
\usepackage{amssymb}
\usepackage{graphicx}
\usepackage[unicode=true,
 bookmarks=false,
 breaklinks=false,pdfborder={0 0 1},backref=section,colorlinks=false]
 {hyperref}

\makeatletter

\providecommand{\tabularnewline}{\\}
\newcommand{\lyxdot}{.}

\floatstyle{ruled}
\newfloat{algorithm}{tbp}{loa}
\providecommand{\algorithmname}{Algorithm}
\floatname{algorithm}{\protect\algorithmname}

\usepackage{hyperref}
\usepackage{microtype}
\usepackage{graphicx} 
\usepackage{array}
\usepackage{amsbsy}
\usepackage{epstopdf}
\usepackage{colortbl} 
\usepackage{caption}
\definecolor{shadedRowColor}{rgb}{0.74,0.88,0.91}
\definecolor{lightgray}{gray}{0.9}

\journal{XXX}

\makeatother

\begin{document}
\begin{frontmatter}

\title{Collaborative filtering via sparse Markov random fields }

\author{Truyen Tran$^{*}$, Dinh Phung, Svetha Venkatesh}

\address{Center for Pattern Recognition and Data Analytics, Deakin University,
Geelong, Victoria, Australia\\
  $^{*}$Corresponding author. E-mail address: truyen.tran@deakin.edu.au }
\begin{abstract}
Recommender systems play a central role in providing individualized
access to information and services. This paper focuses on collaborative
filtering, an approach that exploits the shared structure among mind-liked
users and similar items. In particular, we focus on a formal probabilistic
framework known as Markov random fields (MRF). We address the open
problem of structure learning and introduce a sparsity-inducing algorithm
to automatically estimate the interaction structures between users
and between items. Item-item and user-user correlation networks are
obtained as a by-product. Large-scale experiments on movie recommendation
and date matching datasets demonstrate the power of the proposed method.\end{abstract}
\end{frontmatter}


\global\long\def\BigO{\mathcal{O}}
\global\long\def\U{\mathcal{U}}
\global\long\def\I{\mathcal{I}}
\global\long\def\M{\mathcal{M}}
\global\long\def\obs{\mathbf{o}}
\global\long\def\thetab{\boldsymbol{\theta}}
\global\long\def\pb{\boldsymbol{p}}
\global\long\def\cb{\boldsymbol{c}}
\global\long\def\xb{\boldsymbol{x}}

\global\long\def\A{\mathcal{A}}
\global\long\def\B{\mathcal{B}}
\global\long\def\Loss{\mathcal{L}}
\global\long\def\userset{\mathcal{U}}
\global\long\def\itemset{\mathcal{I}}
\global\long\def\ratemat{\mathcal{R}}

\global\long\def\rateset{\mathcal{S}}
\global\long\def\h{h}
\global\long\def\r{r}
\global\long\def\d{d}
\global\long\def\hb{\boldsymbol{h}}
\global\long\def\rb{\boldsymbol{r}}
\global\long\def\wh{\alpha}
\global\long\def\qb{\boldsymbol{q}}
\global\long\def\pb{\boldsymbol{p}}

\global\long\def\whr{\gamma}
\global\long\def\wr{\beta}
\global\long\def\wrr{\lambda}
\global\long\def\Id{\mathbb{I}}
\global\long\def\Real{\mathbb{R}}

\section{Introduction}

Learning to recommend is powerful. It offers targeted access to information
and services without requiring users to formulate explicit queries.
As the recommender system observes the users, it gradually acquires
users tastes and preferences to make recommendation. Yet its recommendation
can be accurate and sometimes surprising. Recommender systems are
now pervasive at every corner of digital life, offering diverse recommendations
from books \cite{linden2003amazon}, learning courses \cite{farzan2006social},
TV programs \cite{ali2004tivo}, news \cite{das2007gnp}, and many
others (see \cite{lu2015recommender} for an up-to-date survey on
applications). 

An important direction to recommendation is collaborative filtering
(CF). CF is based on the premise that people are interested in common
items, and thus there exists a \emph{shared structure} that enables
transferring one's preference to like-minded users. A highly interpretable
approach is correlation-based, in that our future preference will
be predicted based on either similar users who share the rating history
\cite{resnick94grouplens}, or correlated items that share the previous
raters \cite{sarwar2001ibc}. For example, the popular line ``people
who buy this {[}book{]} also buy ...'' is likely to reflect the correlated
items method.  While this is intuitive, simple correlation methods
might not be effective for several reasons. First, correlation is
heuristic and there is little formal theory that links correlation
to recommendation performance. Second, combining user and item correlations
is desirable but not straightforward. Third, a recommendation should
be equipped with a confidence score, but this is lacking in correlation-based
methods. 

A principled method that addresses these three issues is Preference
Network \cite{Truyen:2007}. A Preference Network is a Markov random
field whose nodes represent preferences by a user on an item, and
edges represent the dependency between items and between users. The
shared structure is encapsulated in the model parameters and enables
future prediction. Model parameters that measure association strength
between items and between users are \emph{jointly} estimated to maximize
the agreement with the data and model prediction. Prediction is based
on the most probable assignment, which comes with quantifiable confidence.

More recent variants and extensions of Preference Network have been
subsequently introduced \cite{defazio2012graphical,gunawardana2008tied,liu2014ordinal,li2015Preference,zou2013iterative}.
However, one important problem still remains, that is the to estimate
the model structure automatically from data. Previous work was mainly
based on heuristics that pick an edge if the correlation is beyond
a predefined threshold. To that end, we propose a sparsity-inducing
framework to learn the edges of the Markov random field directly from
data while maximizing the agreement between data and prediction. It
results in a sparse network, where each item (or user) is connected
to only a handful of other items (or users). Thus it is optimal with
respect to the prediction task, and it frees the model designer from
specifying the structure and justifying the choice. With tens of thousands
of users and items, our MRFs -- with hundreds of millions of free
parameters -- are among the largest MRFs ever studied. With such a
scale, we show how learning is possible using ordinary computers.

We study the capacity of the proposed framework on two online applications:
movie recommendation and match making. In movie recommendation, users
provide ratings for each movie they have watched, and the task is
to predict rating for unseen movies. Likewise in match making, each
user rates a number of profiles of other users, and the recommendation
is to predict how much the user likes new profiles. The movie dataset
is MovieLens 1M with $1$ million ratings by nearly $6$ thousand
users on approximately $4$ thousand movies. The match making dataset
is Dating Agency with 17 million ratings by $135$ thousand users
over $169$ thousand profiles. We show that the MRF-based framework
outperforms well-studied baselines in various scenarios.

To summary, our main contribution is a framework for learning structures
of Markov random fields for collaborative filtering. A by product
of our structure learning framework are item and user correlation
networks, which are useful for further analysis. This extends our
previous work on Preference Network \cite{Truyen:2007}, both in theory
and applications (using new datasets with several orders of magnitude
larger). The rest of the paper is organized as follows. Sec.~\ref{sec:Related-Work}
reviews related work. Sec.~\ref{sec:Structure-learning} presents
our contributions in parameterizations and structure learning. The
proposed frameworks are evaluated extensively in Sec.~\ref{sec:Experimental-Results}.
Sec.~\ref{sec:Conclusion} concludes the paper.

\section{Background \label{sec:Related-Work}}

This section reviews existing work in collaborative filtering (CF)
in general, and presents a detailed account on Markov random fields
for CF in particular.

\subsection{Collaborative filtering}

Recommender systems offer fast personalized access to products and
services and have found applications in numerous places \cite{lu2015recommender,martinez2015model,tejeda2015refore}.
The collaborative filtering approach to recommender systems is based
on the idea that personal preferences can be collaboratively determined
by mind-liked users. In a typical \emph{explicit setting}, a recommender
system maintains a rating database by a set of existing users on a
set of available items. The database can be represented as a sparse
rating matrix, where typically only less than few percents of the
cells are filled. New recommendations will be made for each user for
unseen items without the need of issuing an explicit query. The most
common task is to predict the rating for unseen items, or equivalently,
filling the empty cells in the rating matrix. For that reason, the
task is sometimes referred to as \emph{matrix completion}. In implicit
settings, preferences are not given (e.g., clicks and music streaming).
In this paper, we focus mainly on explicit settings.

The earliest and still most popular methods are \emph{dependency-based},
i.e., co-rated items are correlated and co-raters are interdependent.
The usual technique is $k$-nearest neighbors tailored to collaborative
filtering, which can be user-based or item-based. The \emph{user-based}
method posits that a preference can be borrowed from like-minded users
\cite{resnick94grouplens}. For example, the rating $r_{ui}$ by user
$u$ on item $i$ is predicted as

\begin{eqnarray}
r_{ui} & = & \bar{r}_{u}+\frac{\sum_{v\in U(i)}s(u,v)(r_{vi}-\bar{r}_{v})}{\sum_{v\in U(i)}\left|s(u,v)\right|}\label{eq:user-user}
\end{eqnarray}
where $s(u,v)$ is the correlation between user $u$ and user $v$,
$U(i)$ is the set of all users who rated item $i$, and $\bar{r}_{u}$
is the average rating by user $u$. Note that $s(u,v)$ can be negative,
i.e., two users have opposite tastes. The \emph{item-based} method
predicts rating for a new item based on ratings of other similar items
that the user has rated \cite{sarwar2001ibc}. This is identical to
the user-based method but with the roles of user and item swapped.
The two similarity methods suggest a hybrid that fuses the two predictions
\cite{wang2006unifying}. The computation of the similarity is critical
to the success of the approach. The most common measure is Pearson's
correlation. The main drawback of nearest neighbor is lack of theoretical
justification of the choice of similarity measures and the computation
of rating. 

A more formal method is dependency networks \cite{heckerman2001dni}
which provide a probabilistic interpretation. However, dependency
networks do not offer a consistent probability measure across predictions,
thus limiting its fusion capacity. Markov random fields (MRFs) eliminate
this problem. The first MRF-based recommender system was introduced
in \cite{Truyen:2007} on which the present paper is extended. Factor-graphs,
as an alternative representation of MRF \cite{zou2013iterative},
have been introduced for collaborative filtering but no learning was
done. Rather, the MRF is merely a smoothing mechanism. Learnt MRFs
were investigated in \cite{defazio2012graphical,gunawardana2008tied}
but these are an user-specific version of \cite{Truyen:2007}. More
recently, \cite{liu2014ordinal} extends \cite{Truyen:2007} to incorporate
matrix factorization, but it is still limited to user-specific MRFs. 

\emph{Latent aspects} represent another major approach to collaborative
filtering. Examples include matrix factorization \cite{salakhutdinov2008probabilistic},
RBM \cite{Salakhutdinov-et-alICML07,Truyen:2009a}, PLSA \cite{hofmann2004lsm}
and LDA \cite{marlin2004mur}. These methods assume a low dimensional
representation of rating data, which, once learnt, can be used to
generate unseen ratings. There are evidences suggesting that the dependency-based
and latent aspects approaches are complementary \cite{Truyen:2009a,koren2010factor}.

\subsection{Markov random fields}

\begin{figure}
\begin{centering}
\includegraphics[width=0.5\textwidth]{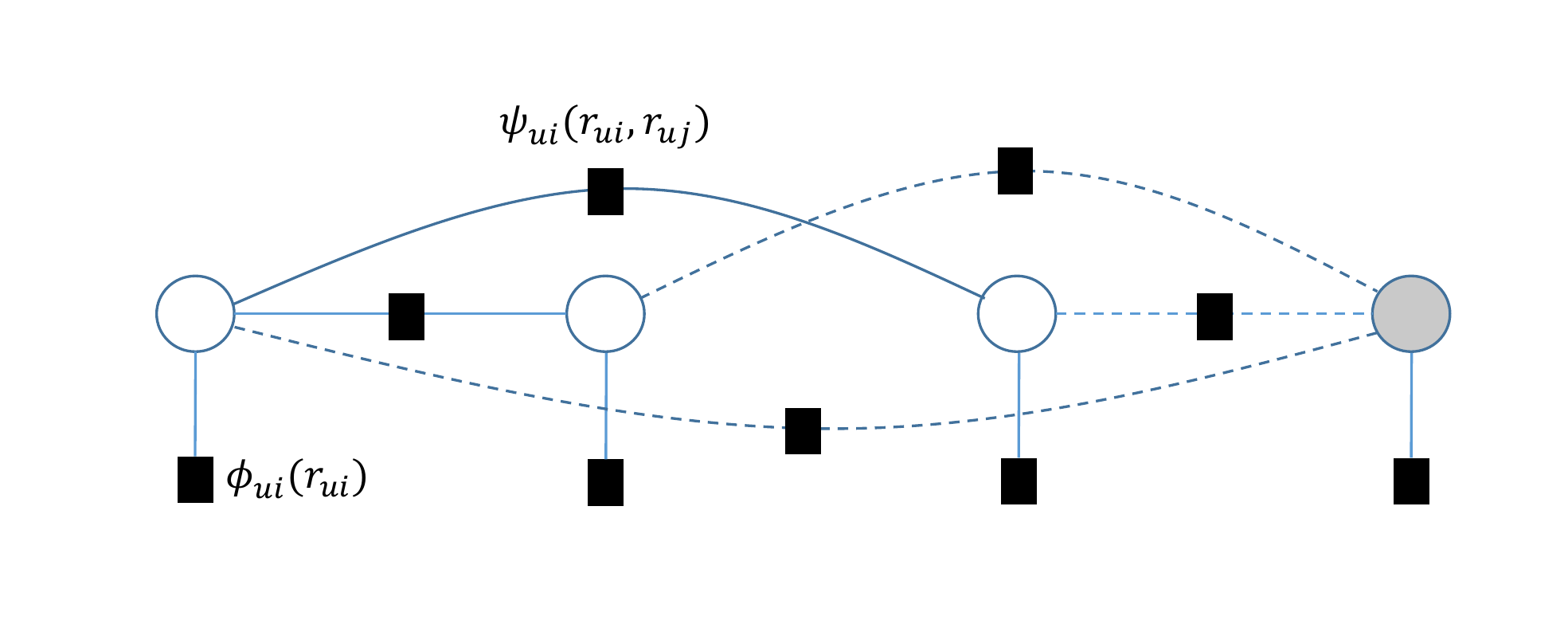}
\par\end{centering}

\protect\caption{Markov random field for a user, represented as a factor graph. Round
nodes represent ratings, filled squares represent potentials (also
known as factors), and edges represent pairwise interaction. New rating
is represented as a shaded node. Dashed lines indicate new edges that
are added at test time and borrowed from other users. \label{fig:Markov-random-field}}
\end{figure}

Markov random field (MRF) is a graph connecting random variables.
A graph is defined as $G=\left(V,E\right)$, where $V$ is the set
of nodes and $E$ is the set of edges (see Fig.~\ref{fig:Markov-random-field}).
Each node represents a random variable. For our purpose, the random
variables $\xb$ are discrete, and we assume that their values are
drawn from a set $\left\{ 1,2,...,K\right\} $. The contribution of
each variable is encoded in a positive function called singleton potential
(or factor) $\phi_{i}(x_{i})$. An edge connecting two variables specifies
a direct association between the two variables, and the association
strength is quantified by a nonnegative pairwise potential $\phi_{ij}(x_{i},x_{j})$
for variable pair $(x_{i},x_{j})$. Note that the absence of an edge
does not rule out the higher-order dependency between two variables.

The joint distribution for all variables is then defined as:
\[
P(\xb)=\frac{1}{Z}\exp\left(-E(\xb)\right)
\]
where $Z$ is the normalizing constant to ensure that $\sum_{\xb}P(\xb)=1$
and $E(\xb)$ is \emph{model energy}, defined as
\[
E(\xb)=-\left(\sum_{i\in V}\log\phi_{i}(x_{i})+\sum_{(i,j)\in E}\log\psi_{ij}(x_{i},x_{j})\right)
\]
A low energy implies high probability, which translates to high compatibility
between variable assignments.

The Hammersley-Clifford theorem \cite{Hammersley-Clifford71} asserts
that, given variable assignments of one's neighborhood, the local
probability is independent of all other variables:
\begin{eqnarray}
P\left(x_{i}\mid\xb_{\neg i}\right) & = & P\left(x_{i}\mid\xb_{N(i)}\right)\propto\exp\left(-E\left(x_{i},\xb_{N(i)}\right)\right)\label{eq:Hammersley-Clifford}
\end{eqnarray}
where $\xb_{\neg i}$ denotes all variables except for $x_{i}$, $N(i)$
is the set of nodes connected to $i$, and 
\[
E\left(x_{i},\xb_{N(i)}\right)=-\left(\log\phi_{i}(x_{i})+\sum_{j\in N(i)}\log\psi_{ij}(x_{i},x_{j})\right)
\]

The neighborhood $N(i)$ is also known as the \emph{Markov blanket}.
This theorem is important because $P\left(x_{i}\mid\xb_{N(i)}\right)$
costs only $K$ time to compute, whereas $P(\xb)$ cannot be evaluated
in polynomial time. Many approximate computations will rely on this
property.

\subsection{MRF for a user (or an item)\label{sub:User-specific-MRF}}

Recall that in the neighborhood-based approach, the correlation between
users (or items) must be estimated, e.g., the $s(u,v)$ in Eq.~(\ref{eq:user-user}).
Ideally the estimation for all user pairs should be directly related
to the final performance measure. A second requirement is that when
making a prediction, we should be able to quantify the confidence
of the prediction. Third, there should be an effective way to combine
user-based and item-based methods. Markov random fields offer a principled
method to meet all three criteria.

Let us start with a MRF per user \cite{Truyen:2007} and then move
to joint MRF for all users in Sec.~\ref{sub:Joint-MRF}. Here the
ratings $\rb=(r_{1},r_{2},...)$ by the user play the role of random
variables. The key is to observe that items rated by the same user
tend to correlate as they reflect user's tastes. Thus each user is
represented by a graph $G=\left(V,E\right)$, where $V$ is the set
of items rated by the user and $E$ is the set of edges connecting
those related items. Each node in $V$ represents a rating variable
$r_{ui}$. A graphical illustration of the MRF is given in Fig.~\ref{fig:Markov-random-field}.

Let $\phi_{ui}(r_{ui})$ be potential function that measures the compatibility
of the rating $r_{ui}$ with user $u$ and item $i$, and $\psi_{ij}(r_{ui},r_{uj})$
encodes the pairwise relationship between two items $(i,j)$. The
model energy is: 
\begin{equation}
E(\rb)=-\left(\sum_{i\in V}\log\phi_{ui}(r_{ui})+\sum_{(i,j)\in E}\log\psi_{ij}(r_{ui},r_{uj})\right)\label{eq:MRF-energy-user}
\end{equation}
A low energy signifies a high compatibility between item-user, and
between item-item. The \emph{local predictive distribution}, following
Eq.~(\ref{eq:Hammersley-Clifford}), is:
\begin{eqnarray}
P\left(r_{ui}\mid\rb_{\neg ui}\right) & \propto & \exp\left(-E\left(r_{ui},\rb_{N(i)}\right)\right)\label{eq:cond-prob}
\end{eqnarray}
where
\begin{equation}
E\left(r_{ui},\rb_{N(i)}\right)=-\left(\log\phi_{ui}(r_{ui})+\sum_{j\in N(i)}\log\psi_{ij}(r_{ui},r_{uj})\right)\label{eq:local-energy}
\end{equation}

As each user only rates a handful of items, it is more efficient to
model only the items each user has rated. Thus the MRFs for all users
will be of different sizes and incorporate different item sets. For
the entire system to make sense, all MRFs must relate in some way.
The key here is that all user-specific models share the same set of
parameters. This parameter sharing enables prediction for unseen ratings,
as we present hereafter.

\subsubsection{Rate prediction}

A fully-specified MRF enables rate prediction of unseen item for reach
user $u$. It is natural to suggest that the best rating will be the
most probable among all ratings, \emph{conditioned on existing ratings},
i.e.:
\begin{eqnarray}
r_{uj}^{*} & =\arg\max_{r_{uj}}P\left(r_{uj}\mid\rb\right)= & \arg\max_{r_{uj}}P\left(r_{uj}\mid\rb_{N(j)}\right)=\arg\min_{r_{uj}}E\left(r_{uj},\rb_{N(j)}\right)\label{eq:rate-pred}
\end{eqnarray}
where $N(j)$ is the set of seen items that are connected to $j$
and $E\left(r_{uj},\rb_{N(j)}\right)$ is the local energy computed
as in Eq.~(\ref{eq:local-energy}). This is a direct application
of the Hammersley-Clifford theorem. A MRF not only can predict new
rating $r_{uj}^{*}$, it also provides the confidence in the prediction
through $P\left(r_{uj}^{*}\mid\rb_{N(j)}\right)$. This property also
enables predicting an \emph{expected rating}:
\begin{equation}
\bar{r}_{uj}=\sum_{k=1}^{K}P\left(r_{uj}=k\mid\rb_{N(j)}\right)k\label{eq:MRF-rate-expect}
\end{equation}
where all probabilities are taken into account.

\subsubsection{Learning}

Learning is to estimate parameters of the potentials $\phi_{ui}(r_{ui})$
and $\psi_{ij}(r_{ui},r_{uj})$. We aim to minimize the disagreement
between the data and the model prediction, typically through the likelihood
function $P(\rb)$. However, estimating the likelihood is generally
intractable due to the exponentially large space of all possible rating
assignments. In our previous work \cite{Truyen:2007}, the negative
log pseudo-likelihood \cite{Besag-74} loss was minimized:
\begin{equation}
\mathcal{L}_{PL}=-\sum_{i\in V}\log P\left(r_{ui}\mid\rb_{\neg i}\right)\label{eq:pseudo-likelihood}
\end{equation}
where $P\left(r_{ui}\mid\rb_{\neg ui}\right)$ is defined in Eq.~(\ref{eq:cond-prob}).
While this loss function is only an approximation to the full negative
log-likelihood, it is appealing because it has the same functional
form as the predictive distribution used in rating prediction of Eq.~(\ref{eq:rate-pred}).

\subsubsection{MRF for an item}

Analogous to the case of building a MRF for a user, we can also build
a MRF for an item. This is because one can rotate the rating matrix
and swap the roles of users and items. Under this view, users are
now naturally dependent under item-based models. In particular, the
model energy in Eq.~(\ref{eq:MRF-energy-user}) can be rewritten
for each item $i$ as
\begin{equation}
E(\rb)=-\left(\sum_{u\in V}\log\phi_{ui}(r_{ui})+\sum_{(u,v)\in E}\log\varphi_{uv}(r_{ui},r_{vi})\right)\label{eq:MRF-energy-item}
\end{equation}
This poses a question to integrate the user-based and item-based views,
which we present next.

\subsection{MRF for entire rating database \label{sub:Joint-MRF}}

\begin{figure}
\begin{centering}
\includegraphics[width=0.5\textwidth]{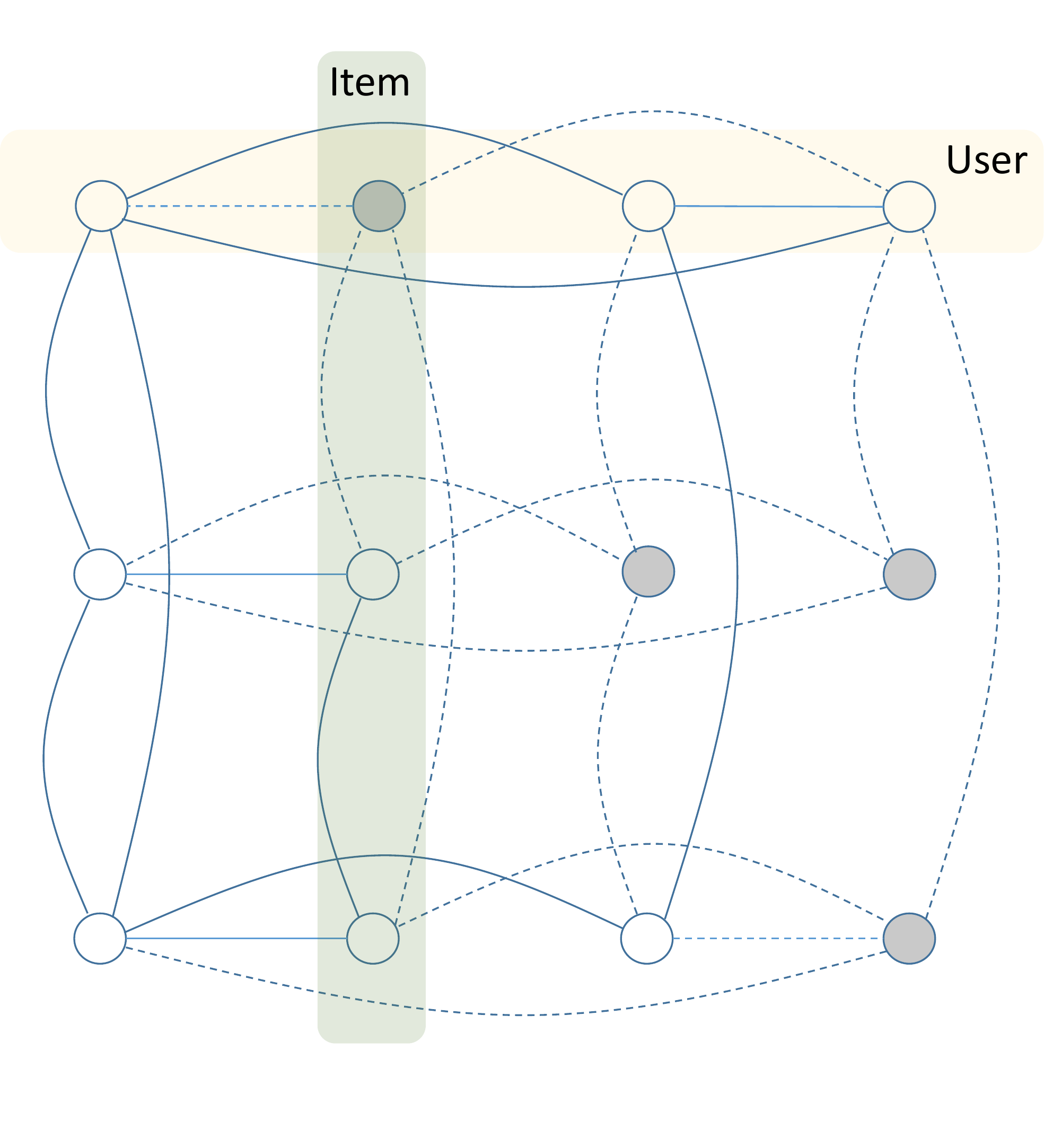}
\par\end{centering}

\protect\caption{Markov random field for full rating database modeling. Factor nodes
in Fig.~\ref{fig:Markov-random-field} are dropped for clarity. When
new ratings are predicted (as shade nodes), their edges are borrowed
from other users and items. Note that we predict one rating at a time,
so there is no direct edges between unseen ratings. \label{fig:full-joint-model}}
\end{figure}

The user-specific and item-specific models presented in the previous
subsections are built upon an unstated assumption that users or items
are drawn randomly and independently from the population. However,
this assumption is unrealistic. First the set of items that an user
rates is subject to availability at the time of rating, and the availability
is evident for all other users. Second, users can be influenced by
other users either implicitly (e.g., an item gets noticed due to its
popularity) or explicitly (e.g., through social contacts). Thus users'
choices are not entirely independent.

We build a single large MRF by joining all user-based and item-based
MRFs together, as illustrated in Fig.~\ref{fig:full-joint-model}.
Denote by $\ratemat$ the \emph{entire rating database}, the full
joint model energy is: 

\begin{eqnarray*}
E(\ratemat) & = & -\left(\sum_{r_{ij}\in\ratemat}\log\phi_{ui}(r_{ui})+\sum_{u}\sum_{(i,j)\in E(u)}\log\psi_{ij}(r_{ui},r_{uj})+\sum_{i}\sum_{(u,v)\in E(i)}\log\varphi_{uv}(r_{ui},r_{vi})\right)
\end{eqnarray*}
where $E(u)$ and $E(i)$ are the set of edges specific to user $u$
and item $i$, respectively. Applying the Hammersley-Clifford theorem
in Eq.~(\ref{eq:Hammersley-Clifford}), the local predictive distribution
becomes:
\begin{eqnarray*}
P\left(r_{ui}\mid\ratemat_{\neg ui}\right) & \propto & \exp\left(-E\left(r_{ui},\boldsymbol{R}_{N(u,i)}\right)\right)
\end{eqnarray*}
where $\ratemat_{\neg ui}$ is all ratings except for $r_{ui}$, $N(u,i)$
is the set of neighbors of the pair $(u,i)$ (e.g., see the shaded
row and column in Fig.~\ref{fig:full-joint-model}), and 
\begin{eqnarray}
E\left(r_{ui},\boldsymbol{R}_{N(u,i)}\right) & = & -\left(\log\phi_{ui}(r_{ui})+\sum_{(u,j)\in N(u,i)}\log\psi_{ij}(r_{ui},r_{uj})+\sum_{(v,i)\in N(u,i)}\log\varphi_{uv}(r_{ui},r_{vi})\right)\label{eq:local-energy-joint}
\end{eqnarray}

\subsubsection{Rate prediction}

Similar to the case of separate MRFs in Sec.~\ref{sub:User-specific-MRF},
rate prediction for a new user/item pair is based on the existing
ratings associated with the user and the item as follows:

\begin{eqnarray}
r_{vj}^{*} & = & \arg\max_{r_{vj}}P\left(r_{vj}\mid\boldsymbol{R}_{N(v,j)}\right)=\arg\min_{r_{vj}}E\left(r_{vj},\boldsymbol{R}_{N(v,j)}\right)\label{eq:MRF-rate-predict-joint}
\end{eqnarray}
where $E\left(r_{vj},\boldsymbol{R}_{N(u,j)}\right)$ is local energy
computed as in Eq.~(\ref{eq:local-energy-joint}). The \emph{expected
rating} can be computed as:
\begin{equation}
\bar{r}_{vj}=\sum_{k=1}^{K}P\left(r_{vj}\mid\boldsymbol{R}_{N(u,j)}\right)k\label{eq:MRF-rate-expect-joint}
\end{equation}

\section{Structure learning \label{sec:Structure-learning}}

The MRFs presented in Sec.~\ref{sub:User-specific-MRF} require the
model structures to be pre-determined by hand. In this section, we
present a method to learn the structure from data. We first introduce
several parameterization schemes of the potentials (Sec.~\ref{sub:Parameterizations})
that facilitate structure learning (Sec.~\ref{sub:Structure-learning}).

\subsection{Log-linear parameterizations \label{sub:Parameterizations}}

It is important to note that ratings are not merely discrete but also
ordinal. That is, if the true rating is $3$, it is better to get
closer to $3$ in prediction (e.g., $2$ and $4$ rather than $1$
and $5$). This is unlike unordered discrete variables, where all
the options are \emph{a priori} equal. We present here three log-linear
parameterization schemes to capture this ordinal property: \emph{linear-by-linear},
\emph{Gaussian}, and \emph{smoothness}. In what follows, we will assume
that there are $N$ users and $M$ items in the rating database whose
values are drawn from the ordered set $\left\{ 1,2,..,K\right\} $.

\subsubsection{Linear-by-linear parameterization \label{sub:Linear-by-linear-parameterization}}

The first scheme is a linear-by-linear parameterization in which the
potential functions in Eq.~(\ref{eq:MRF-energy-user}) have the following
forms:
\begin{eqnarray}
\phi_{ui}(r_{ui}=k) & = & \exp\left(\alpha_{ik}+\beta_{uk}\right)\label{eq:singleton-poten-multinomial}\\
\psi_{ij}(r_{ui},r_{uj}) & = & \exp\left(\omega_{ij}\left(r_{ui}-\bar{r}_{i}\right)\left(r_{uj}-\bar{r}_{j}\right)\right)\label{eq:pairwise-poten-multinomial-item-item}\\
\varphi_{uv}(r_{ui},r_{vi}) & = & \exp\left(w_{uv}\left(r_{ui}-\bar{r}_{u}\right)\left(r_{vi}-\bar{r}_{v}\right)\right)\label{eq:pairwise-poten-multinomial-user-user}
\end{eqnarray}
where $\left\{ \alpha_{ik},\beta_{uk}\right\} $ are rating biases
for item $i$ and user $u$, respectively; $\left\{ \omega_{ij},w_{uv}\right\} $
are pairwise interaction parameters for item pair $(i,j)$ and user
pair $(u,v)$, respectively; and $\left\{ \bar{r}_{i},\bar{r}_{u}\right\} $
are mean rates for item $i$ and user $u$, respectively. The bias
$\alpha_{ik}$ reflects the overall quality of an item, regardless
of the user. For example, popular movies tend to receive higher ratings
than average. The bias $\beta_{uk}$ indicates the tendency that a
user chooses a particular rating. This is because some users may or
may not be critical in their rating, and some users only rate items
that they like, ignoring those they do not like. 

The pairwise potential $\psi_{ij}(r_{ui},r_{uj})$ reflects the ordering
of both $r_{i}$ and $r_{j}$. Since its log is linear in either variable,
this parameterization is called linear-by-linear model \cite[Chap. 8]{agresti1990cda}.
This parameterization have $NK+MK+\frac{1}{2}M(M-1)$ parameters.
Similar properties hold for $\varphi_{uv}(r_{ui},r_{vi})$.

\paragraph{Remark }

The pairwise potential can be parameterized differently, e.g., $\psi_{ij}(r_{i}=k_{1},r_{j}=k_{2})=\exp\left(\omega_{ijk_{1}k_{2}}\right)$.
However, since this multiplies the number of parameters by a factor
of $K^{2}$, it is expensive to compute and less reliable to learn.
Thus we do not investigate this option in the paper. The main drawback
of this approach is the treatment of ordinal ratings as categorical,
and thus losing important information.

\subsubsection{Gaussian parameterization \label{sub:Gaussian-parameterization}}

An approximation to ordinal treatment is the Gaussian parameterization
scheme, where ratings are considered as continuous variables. The
potential functions in Eq.~(\ref{eq:MRF-energy-user}) can be specified
as:

\begin{eqnarray}
\phi_{ui}(r_{ui}) & = & \exp\left(-\frac{\left(r_{ui}-\alpha_{i}-\beta_{u}\right)^{2}}{2}\right)\label{eq:singleton-poten-gauss}\\
\psi_{ij}(r_{ui},r_{uj}) & = & \exp\left(\omega_{ij}r_{ui}r_{uj}\right)\label{eq:pairwise-poten-gauss-item-item}\\
\varphi_{uv}(r_{ui},r_{vi}) & = & \exp\left(w_{uv}r_{ui}r_{vi}\right)\label{eq:pairwise-poten-gauss-user-user}
\end{eqnarray}
Thus $\psi_{ij}(r_{ui},r_{uj})$ captures the linear association between
items, similar to the linear-by-linear parameterization in Sec.~\ref{sub:Linear-by-linear-parameterization}.
This parameterization have $N+M+\frac{1}{2}M(M-1)$ parameters. Similar
properties hold for $\varphi_{uv}(r_{ui},r_{vi})$.

\paragraph{Remark }

The model is log-linear because $\phi_{ui}(r_{ui})\propto\exp\left(0.5\left\{ -r_{ui}^{2}+(\alpha_{i}+\beta_{u})r_{ui}\right\} \right)$
which is log-linear in $\alpha_{i},\beta_{u}$, and $\psi_{ij}(r_{ui},r_{uj})$
and $\varphi_{uv}(r_{ui},r_{vi})$ are log-linear in $\omega_{ij}$
and $w_{uv}$, respectively.

The local predictive distribution has the following form: 
\begin{eqnarray*}
P\left(r_{uj}\mid\rb_{N(j)}\right) & \propto & \exp\left(-\frac{\left(r_{j}-\alpha_{j}-\sum_{i\in N(j)}\omega_{ij}r_{ui}\right)^{2}}{2}\right)
\end{eqnarray*}
When $r_{uj}$ is allowed to take value in entire $\mathbb{R}$, it
is essentially a normal distribution of mean $\alpha_{j}+\sum_{i\in N(j)}\omega_{ij}r_{ui}$.

\paragraph*{Rate normalization}

The Gaussian model assumes ratings of variance $1$. It necessitates
normalization before training can start. Our normalization is a two-step
procedure:
\begin{enumerate}
\item The first step normalizes data per user. Adapting from \cite{hofmann2004lsm},
we transform the rating as follows
\[
\hat{r}_{ui}\leftarrow\frac{r_{ui}-\bar{r}_{u}}{\bar{s}_{u}},
\]
where $\bar{r}_{u}$ and $\bar{s}_{u}$ are the mean rating and smoothed
deviation by user $u$, respectively. The smoothed deviation is estimated
from the deviation $s_{u}$ as follows:
\[
\bar{s}_{u}=\sqrt{\frac{5s^{2}+n_{u}s_{u}^{2}}{5+n_{u}}},
\]
where $s$ is the global deviation for the whole train data, $n_{u}$
is the number of items rated by user $u$. Thus $\bar{s}_{u}$ is
between $s$ and $s_{u}$ -- $\bar{s}_{u}$ is closer to $s$ if $n_{u}$
is small, and to $s$ otherwise.
\item The second step normalizes data per item.
\[
\hat{\hat{r}}_{ui}\leftarrow\frac{\hat{r}_{ui}-\bar{\hat{r}}_{i}}{\bar{\hat{s}}_{i}},
\]
where $\bar{\hat{r}}_{i}$ is the mean rate for item $i$ after the
first step and $\bar{\hat{s}}_{i}$ is the smoothed deviation computed
as: 
\[
\bar{\hat{s}}_{i}=\sqrt{\frac{5+\sum_{u\in U(i)}\left(\hat{r}_{ui}-\bar{\hat{r}}_{i}\right)}{5+m_{i}}}.
\]
Thus $\bar{\hat{s}}_{i}$ is closer to $1$ if $m_{i}$ is small.
\end{enumerate}
At prediction time, the reverse process is performed to recover the
original scale:
\[
r_{ui}\leftarrow\bar{r}_{u}+\bar{s}_{u}\left(\bar{\hat{r}}_{i}+\bar{\hat{s}}_{i}\hat{\hat{r}}_{ui}\right)
\]

\subsubsection{Smoothness parameterization \label{sub:Smoothness-parameterization}}

While Gaussian parameterization respects the ordinal property of ratings,
the Gaussian assumption could be too strong. In this approach we employ
an ordinal parameterization following \cite{Truyen:2007,Truyen:2009a}:

\begin{eqnarray}
\phi_{ui}(r_{ui}=k_{1}) & = & \exp\left(-\sum_{k_{2}=1}^{K}\left(\alpha_{ik_{2}}+\beta_{uk_{2}}\right)\left|k_{1}-k_{2}\right|\right)\label{eq:singleton-poten-ord}\\
\psi_{ij}(r_{ui},r_{uj}) & = & \exp\left(-\omega_{ij}\left|r_{ui}-r_{uj}\right|\right)\label{eq:pairwise-poten-ord-item-item}\\
\varphi_{uv}(r_{ui},r_{vi}) & = & \exp\left(-w_{uv}\left|r_{ui}-r_{vi}\right|\right)\label{eq:pairwise-poten-ord-user-user}
\end{eqnarray}
This parameterization have $NK+MK+\frac{1}{2}M(M-1)$ parameters.
The singleton potential $\phi_{ui}(r_{ui})$ captures the relative
distances of the current rating from anchor points $\left\{ 1,2,..,K\right\} $.
The pairwise potential $\psi_{ij}(r_{ui},r_{uj})$ enables smoothness
between neighbor ratings by the same user, parameterized by $\omega_{ij}$.
A similar property holds for $\varphi_{uv}(r_{ui},r_{vi})$ for the
same item.

\subsection{Structure learning for user-specific models \label{sub:Structure-learning}}

Given our log-linear parameterizations, an edge contributes to model
energy only if its parameter is non-zero. Thus, structure learning
reduces to estimating non-zero pairwise parameters $\left\{ \omega_{ij}\right\} $
and $\left\{ w_{uv}\right\} $. For clarity, let us start with user-specific
models (Sec.~\ref{sub:User-specific-MRF}). We propose to minimize
the following $\ell_{1}$-penalized loss:

\begin{equation}
\thetab^{*}=\arg\min_{\thetab}\left(\mathcal{L}(\thetab)+\lambda_{1}\sum_{i}\sum_{j>i}\left|\omega_{ij}\right|\right)\label{eq:param-learn}
\end{equation}
where $\thetab=\left(\alpha,\beta,\omega\right)$ denotes the set
of model parameters, $\mathcal{L}(\thetab)$ is the loss function,
and $\lambda_{1}>0$ is the regularization parameter. This setting
has the following properties:
\begin{itemize}
\item The $\ell_{1}$-penalty drives weakly weights of correlated item pairs
towards zero, thus achieving a sparse item graph solution as a by-product,
and
\item The hyper-parameter $\lambda_{1}$ controls the sparsity of the solution,
that is, the higher $\lambda_{1}$ leads to more sparse solutions.
\end{itemize}
A typical loss function is the negative log-likelihood, i.e., $\mathcal{L}(\thetab)=-\log P(\rb;\thetab)$,
where $P(\rb;\thetab)$ is defined in Eq.~(\ref{eq:MRF-energy-user}).
When the gradient of the loss function is available, we can use the
gradient ascent method to optimize it. For example, the gradient descent
update for the pairwise parameters is
\begin{equation}
\omega_{ij}\leftarrow\omega_{ij}-\eta\left(\partial_{\omega_{ij}}\mathcal{L}(\thetab)+\lambda_{1}\mbox{sign}\left(\omega_{ij}\right)\right)\label{eq:param-update}
\end{equation}
where $\eta>0$ is the learning rate. However, since $P(\rb;\thetab)$
and the gradient $\partial_{\omega_{ij}}\mathcal{L}(\thetab)$ are
intractable to compute exactly, we resort to surrogate methods --
one approximates the loss (pseudo-likelihood), the other approximates
the gradient (contrastive divergence). While there are many algorithms
to deal with non-smooth gradient due to the step function $\mbox{sign}()$,
we employ here a simple approximate solution:
\[
\left|\omega_{ij}\right|\approx\sqrt{\epsilon^{2}+\omega_{ij}^{2}}
\]
for $0<\epsilon\ll1$, which has the smooth gradient $\left[\epsilon^{2}+\omega_{ij}^{2}\right]^{-1/2}\omega_{ij}$.

\subsubsection{Pseudo-likelihood (PL) \label{sub:Pseudo-likelihood}}

The pseudo-likelihood loss is defined in Eq.~(\ref{eq:pseudo-likelihood}).
The loss and its gradient can be computed exactly. For example, for
Gaussian parameterization, the gradient for pairwise parameters is
\[
\partial_{\omega_{ij}}\mathcal{L}_{PL}(\thetab)=\mu_{i}r_{uj}+\mu_{j}r_{ui}-2r_{ui}r_{uj}
\]
where
\[
\mu_{i}=\alpha_{i}+\sum_{j_{1}\in N(i)}\omega_{ij_{1}}r_{uj_{1}}
\]

\subsubsection{Contrastive divergence (CD) \label{sub:Contrastive-divergence}}

Alternatively, we use the original loss function, but approximate
its gradient. For example, for Gaussian parameterization, the derivative
with respect to pairwise parameters reads: 
\begin{equation}
\partial_{\omega_{ij}}\mathcal{L}(\thetab)=\mathbb{E}\left[r_{ui}r_{uj}\right]-r_{ui}r_{uj}\label{eq:CD-gradient}
\end{equation}
The expectation $\mathbb{E}\left[r_{ui}r_{uj}\right]$ can be approximated
by samples drawn from the distribution $P(\rb)$ as follows:
\[
\mathbb{E}\left[r_{ui}r_{uj}\right]\approx\frac{1}{n}\sum_{s=1}^{n}r_{ui}^{(s)}r_{uj}^{(s)}
\]
Since full sampling until convergence is expensive, we employ a short-cut
called $c$-step contrastive divergence \cite{Hinton02} (CD). More
specifically, we start a Markov chain from the data $\rb$ itself,
and apply Gibbs sampling for $c$ scans over all items. The Gibbs
sampling iteratively draws a rating at a time using $r_{ui}^{(s)}\sim P\left(r_{ui}\mid\rb_{\neg ui}^{(s)}\right)$,
updating the samples along the way. Typically $c$ is a small number.
The sample at the end of the $c$ scans will be retained to approximate
the gradients.

\subsubsection{Reducing computational complexity \label{sub:Reducing-complexity}}

The prediction complexity per item is $\mathcal{O}(m_{max}K)$ for
the linear-by-linear and Gaussian models and $\mathcal{O}(m_{max}K^{2})$
the smoothness model, where $m_{max}$ is the maximum number of items
per users. Typically, this is fast because $m_{max}\ll M$. 

In learning, we use mini-batches of $b$ users to compute gradients
to speed up. At each parameter update, learning takes $\mathcal{O}\left(\left(m_{max}+m_{max}^{2}\right)bK\right)$
time to compute gradient and $\mathcal{O}\left((N+M)K+\frac{1}{2}M^{^{2}}\right)$
time to update parameters. Memory consumes $\mathcal{O}\left((M+N)K+\frac{1}{2}M^{^{2}}\right)$
space to store parameters. We propose to reduce these demanding memory
and time by specifying the max number of neighbors, e.g., $m\ll M$,
giving a time complexity of $\mathcal{O}\left((M+N)K+Mm\right)$.
One way to specify the neighbor is pre-filtering by correlation measures,
that is, we keep only highly correlated item pairs. However, this
method creates a significant run-time overhead of $\mathcal{O}(\log m)$
by checking if a pair is pre-selected. In this paper, we use a simple
method for fast memory access: each item is connected to $m$ most
popular items.

\subsection{Structure learning of the entire rating database}

Extension to the entire rating database of Sec.~\ref{sub:Joint-MRF}
is straightforward. Eq.~(\ref{eq:param-learn}) is now extended to:

\begin{equation}
\thetab^{*}=\arg\min_{\thetab}\left(\mathcal{L}(\thetab)+\lambda_{1}\sum_{i}\sum_{j>i}\left|\omega_{ij}\right|+\lambda_{2}\sum_{u}\sum_{v>u}\left|w_{uv}\right|\right)\label{eq:param-learn-joint}
\end{equation}
where $\thetab$ now consists of all user-specific and item-specific
parameters, $\omega_{ij}$ is item-item parameter, $w_{uv}$ is the
user-user parameter, and $\lambda_{1},\lambda_{2}>0$.

\begin{algorithm}
Initialize parameters to zeros.

\textbf{Loop} until convergence:

\qquad{}\emph{/{*} Fix the user-user parameters {*}/}

\qquad{}\textbf{For} each user batch

\qquad{}\qquad{}\textbf{For} each user $u$ in the batch

\qquad{}\qquad{}\qquad{}Compute the gradient of $\begin{cases}
\log P\left(\rb_{u}\mid\ratemat_{\neg u}\right) & \mbox{if using contrastive divergence}\\
\sum_{i\in I(u)}\log P\left(r_{ui}\mid\ratemat_{\neg ui}\right) & \mbox{otherwise}
\end{cases}$

\qquad{}\qquad{}\textbf{EndFor}

\qquad{}\qquad{}Update relevant biases $(\alpha,\beta)$ and item-item
pairwise parameters $\omega$

\qquad{}\textbf{EndFor}\\

\qquad{}\emph{/{*} Fix the item-item parameters {*}/}

\qquad{}\textbf{For} each item batch

\qquad{}\qquad{}\textbf{For} each item $i$ in the batch

\qquad{}\qquad{}\qquad{}Compute the gradient of $\begin{cases}
\log P\left(\rb_{i}\mid\ratemat_{\neg i}\right) & \mbox{if using contrastive divergence}\\
\sum_{u\in U(i)}\log P\left(r_{ui}\mid\ratemat_{\neg ui}\right) & \mbox{otherwise}
\end{cases}$

\qquad{}\qquad{}\textbf{EndFor}

\qquad{}\qquad{}Update biases $(\alpha,\beta)$ and user-user pairwise
parameters $w$

\qquad{}\textbf{EndFor}

\textbf{EndLoop}

\protect\caption{Online alternating blockwise pseudo-likelihood for joint learning.\label{alg:Block-pseudo-likelihood}}
\end{algorithm}

We present an efficient algorithm that relies on \emph{blockwise pseudo-likelihood}.
For example, an user's rating set is a block from which the conditional
distribution $P\left(\rb_{u}\mid\ratemat_{\neg u}\right)$ is estimated,
where $\ratemat_{\neg u}$ is the set of all ratings except for those
by user $u$. Likewise, we also have an item block with the conditional
distribution $P\left(\rb_{i}\mid\ratemat_{\neg i}\right)$, where
$\ratemat_{\neg i}$ is the set of all ratings except for those on
item $i$. Thus it is an extension of the standard pointwise pseudo-likelihood
in \cite{Besag-74}.

This suggests an alternating procedure between updating user-based
models and item-based models. For each model, the learning techniques
(pointwise pseudo-likelihood and contrastive divergence) presented
in Sec.~\ref{sub:Structure-learning} are applicable. When computing
$P\left(\rb_{u}\mid\ratemat_{\neg u}\right)$, we need to take the
user neighborhoods of $r_{ui}$ for each item $i$ rated by user $u$.
Likewise, when computing $P\left(\rb_{i}\mid\ratemat_{\neg i}\right)$,
we need to account for the item neighborhoods of $r_{ui}$ for each
user $u$ who rated item $i$. 

The overall algorithm is presented in Alg.~\ref{alg:Block-pseudo-likelihood}.
It is an online-style algorithm for speeding up. In particular, we
update parameters after every small batch of either users or items.

\subsubsection{Reducing computational complexity}

Similar to those described in Sec.~\ref{sub:Reducing-complexity}
we use the method of limiting neighborhood size to $m\ll M$ items
and $n\ll N$ users. The prediction complexity per item is $\mathcal{O}\left((n+m)K\right)$
for linear-by-linear (Sec.~\ref{sub:Linear-by-linear-parameterization})
and Gaussian (Sec.~\ref{sub:Gaussian-parameterization}) parameterizations
and $\mathcal{O}\left((n+m)K^{2}\right)$ for the smoothness parameterization
(Sec.~\ref{sub:Smoothness-parameterization}). At each parameter
update after a batch of size $b$, learning takes $\mathcal{O}\left(\left(n+m+n^{2}+m^{2}\right)bK\right)$
time to compute gradient for linear-by-linear and Gaussian parameterizations
and $\mathcal{O}\left(\left(n+m+n^{2}+m^{2}\right)bK^{2}\right)$
for the smoothness parameterization. Parameter update takes $\mathcal{O}\left((N+M)K+nN+mM\right)$
time for all parameterizations.

\section{Experimental results \label{sec:Experimental-Results}}

In this section, we present a comprehensive evaluation of our proposed
method on two applications: \emph{movie recommendation} and \emph{online
date matching}.

\subsection{Experimental setup}

For rating prediction, we report three measures: \emph{the root-mean
square error} (RMSE), the \emph{mean absolute error} (MAE) and the
\emph{log-likelihood} (LL).

\begin{eqnarray*}
RMSE & = & \sqrt{\frac{1}{Y}\sum_{u=1}^{N}\sum_{j\in J(u)}\left(r_{uj}-\bar{r}_{uj}\right)^{2}}\\
MAE & = & \frac{1}{Y}\sum_{u=1}^{N}\sum_{j\in J(u)}\left|r_{uj}-r_{uj}^{*}\right|\\
LL & = & \frac{1}{Y}\sum_{u=1}^{N}\sum_{j\in J(u)}\log P(r_{uj})
\end{eqnarray*}
where $J(u)$ is the set of new items for user $u$, $\bar{r}_{uj}$
is the expected rating from Eqs.~(\ref{eq:MRF-rate-expect},\ref{eq:MRF-rate-expect-joint}),
$r_{uj}^{*}$ is predicted rating from Eqs.~(\ref{eq:rate-pred},\ref{eq:MRF-rate-predict-joint}),
and $Y$ is total number of new predictions (i.e., $Y=\sum_{u=1}^{N}\sum_{j\in J(u)}1$).
The RMSE and MAE measure the distance from the predicted rating and
the true rating (i.e., the smaller the better). The log-likelihood,
on the hand, measures how the model fits the unseen data (i.e., the
larger the better).

\subsubsection{MRF implementation}

Learning for MRFs is iterative in that for each iteration, first user-specific
models are updated followed by item-specific models. Parameters are
updated after every batch of $b=100$ users or items. Learning rate
$\eta$ is set at 0.1 for biases and 0.01 for pairwise parameters.
To speed up, at the early learning stages, only biases are learnt.
Once the learning starts getting saturated, pairwise parameters are
then introduced. Note that this schedule does not alter the solution
quality since the objective function is concave. But it may improve
the convergence speed because pairwise gradients are much more expensive
to estimate. Rate prediction is by Eq.~(\ref{eq:rate-pred}) when
an integer output is expected (e.g., for estimating MAE), and by Eq.~(\ref{eq:MRF-rate-expect})
for a real-valued output (e.g., for estimating RMSE).

\subsubsection{Baselines\label{sub:Baselines}}

For comparison we implemented $3$ simple baselines, one using user
mean-rating $\bar{r}_{u}$ as prediction for user $u$, another using
item mean-rating $\bar{r}_{i}$ for item $i$, and a weighted mean
accounting for variances:
\begin{equation}
\bar{r}_{ui}=\frac{\bar{r}_{u}/s_{u}+\bar{r}_{i}/s_{i}}{1/s_{u}+1/s_{i}}\label{eq:weighted-mean}
\end{equation}
where $s_{u}$ and $s_{i}$ are rating deviations for user $u$ and
item $i$, respectively.

We also implemented one of the best performing techniques in the Netflix
challenge: regularized singular value decomposition (RSVD), also known
as probabilistic matrix factorization \cite{salakhutdinov2008probabilistic}.
They are both latent space models. The RSVD assumes a generative Gaussian
distribution of ratings: 
\begin{eqnarray*}
r_{ui} & \sim & \mathcal{N}\left(\mu_{ui},\sigma_{ui}^{2}\right)\\
\mu_{ui} & = & a_{i}+b_{u}+\sum_{f=1}^{F}A_{if}B_{uf}\\
\sigma_{ui}^{2} & = & e^{\gamma_{u}+\nu_{i}}\\
A_{if} & \sim & \mathcal{N}\left(0,\lambda^{-1}\right)\\
B_{uf} & \sim & \mathcal{N}\left(0,\lambda^{-1}\right)
\end{eqnarray*}
Here $\left\{ a_{i},b_{u},A_{if},B_{uf},\gamma_{u},\nu_{i}\right\} $
are free parameters and $F$ is the latent dimensions. Unlike existing
RSVD implementation where $\sigma_{ui}^{2}=1$, we also estimate the
variance $\sigma_{ui}^{2}$ for the purpose of estimating a better
data likelihood. The prior variance $\lambda$ is tuned for the best
performance.

\subsubsection{Validation}

For each dataset, we remove those infrequent users who have less than
$30$ ratings. Then $5$ items per user are held out for validation,
$10$ for testing and the rest for training. We make sure that the
prediction is time-sensible, i.e., the training time-stamps precedes
validation which precedes testing. For MRFs, learning is monitored
using the pseudo-likelihood on the validation data. Learning is stopped
if there is no improvement of pseudo-likelihood on the validation
set.

\subsection{Movie recommendation}

For movie recommendation, we use the MovieLens 1M\footnote{http://www.grouplens.org/node/12}
dataset with 1 million ratings in a $5$-star scale given by $6$
thousand users on approximately $4$ thousand movies. After removing
infrequent users, we retain $5.3$ thousand users, $3.3$ thousands
items and $901.1$ thousand ratings. The mean rating is $3.6$ (std:
$1.1$) and the rating matrix is sparse, with only $5.2\%$ cells
filled. On average, a movie is rated $276$ times (median: $145$),
and an user rates $170$ movies (median: $99$).

\subsubsection{Learning curves}

\begin{figure}
\begin{centering}
\begin{tabular}{cc}
\includegraphics[width=0.45\textwidth]{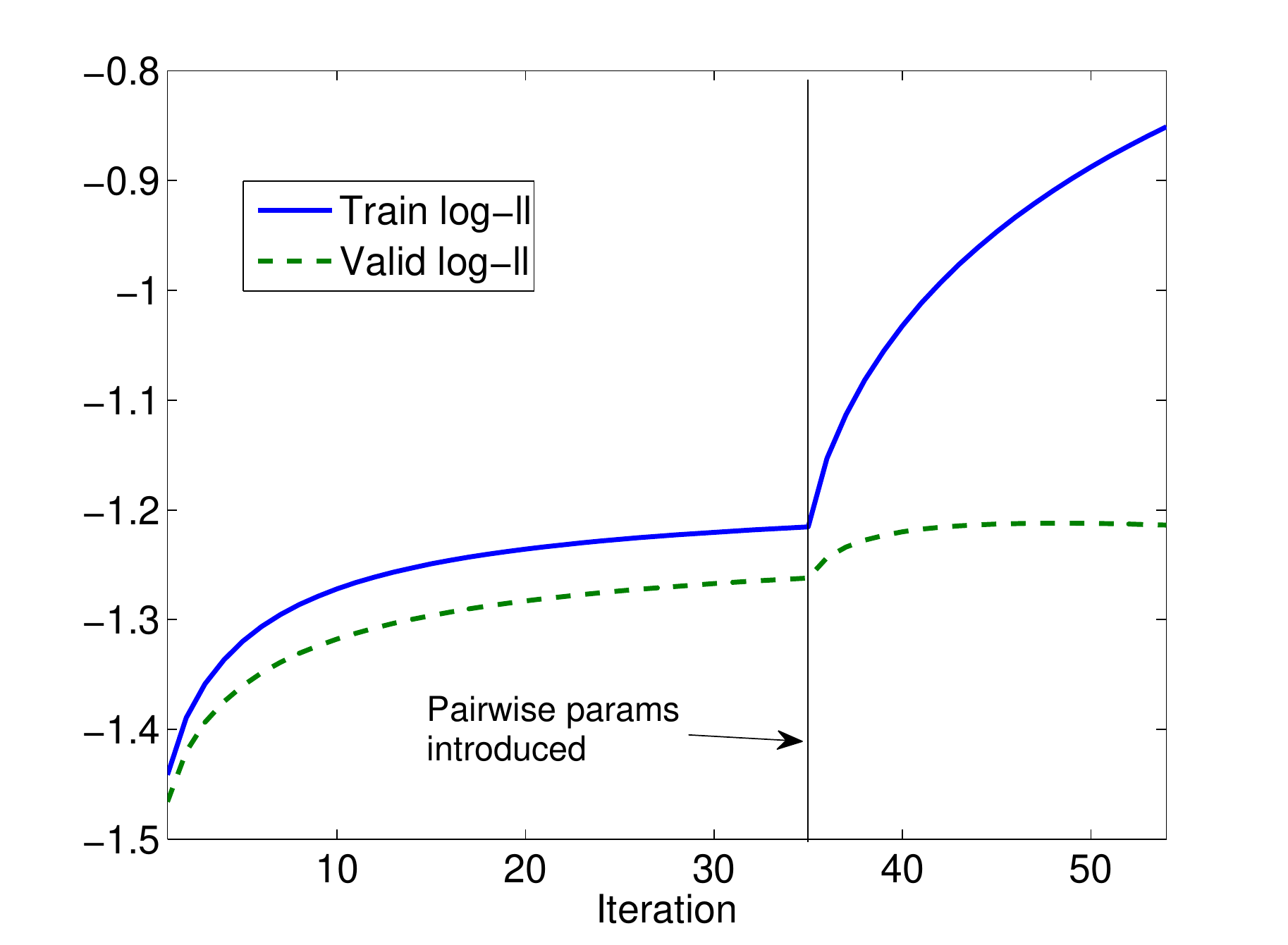} & \includegraphics[width=0.45\textwidth]{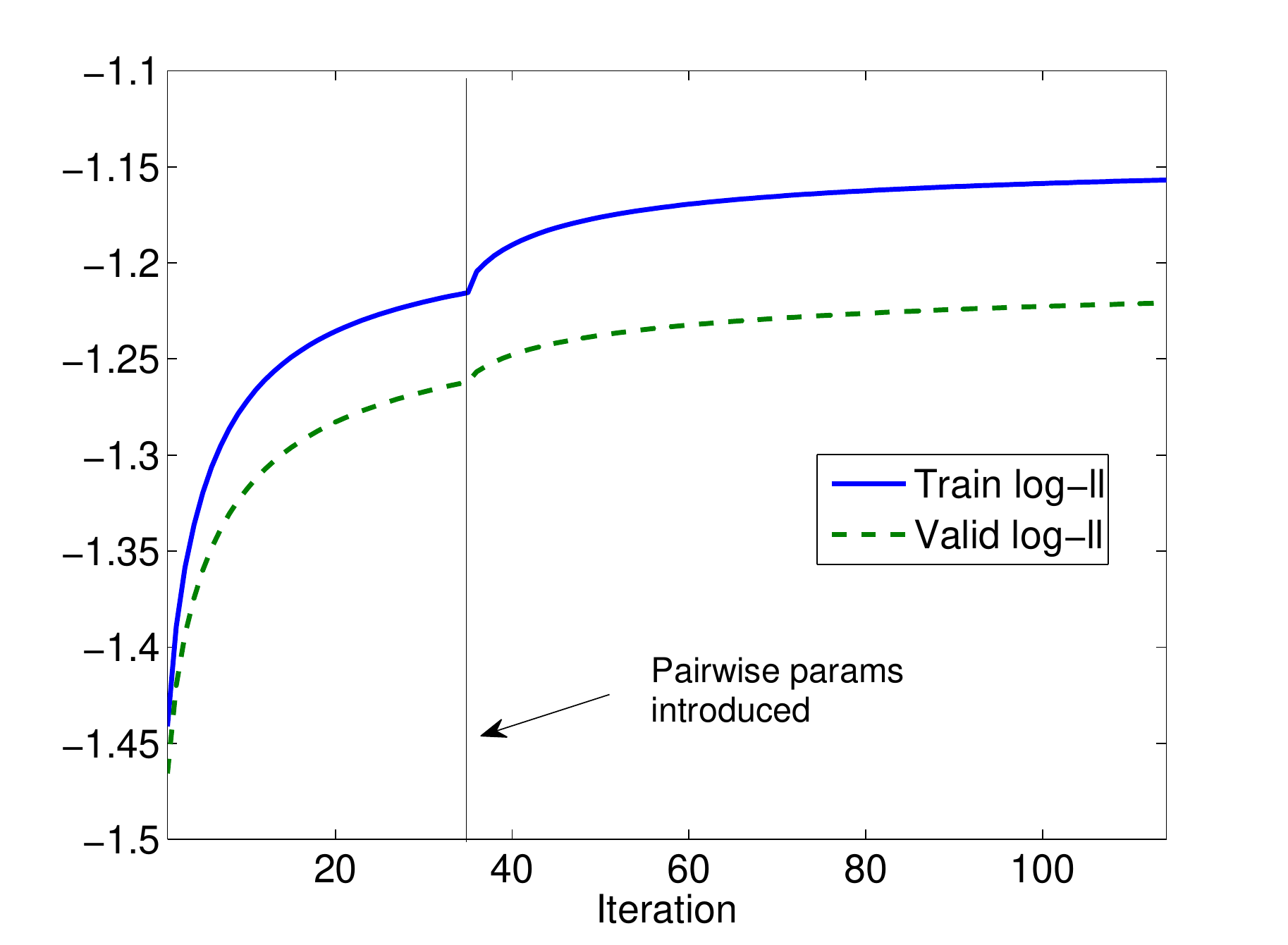}\tabularnewline
$\lambda_{1}=10^{-5}$ & $\lambda_{1}=10^{-2}$\tabularnewline
\end{tabular}
\par\end{centering}

\protect\caption{Learning curves for pseudo-likelihood. Data: MovieLens 1M; model:
user-specific with \emph{smoothness parameterization} (Sec.~\ref{sub:Smoothness-parameterization})
learnt from \emph{pseudo-likelihood} (Sec.~\ref{sub:Pseudo-likelihood}).
There are two stages: first only biases are learnt, then pairwise
parameters are introduced. $\lambda_{1}$ is the sparsity inducing
factor as in Eq.~(\ref{eq:param-learn}). The gap between the training
log-likelihood and the validation indicates potential overfitting.
\label{fig:Learning-curve}}
\end{figure}

\begin{figure}
\begin{centering}
\begin{tabular}{cc}
\includegraphics[width=0.45\textwidth]{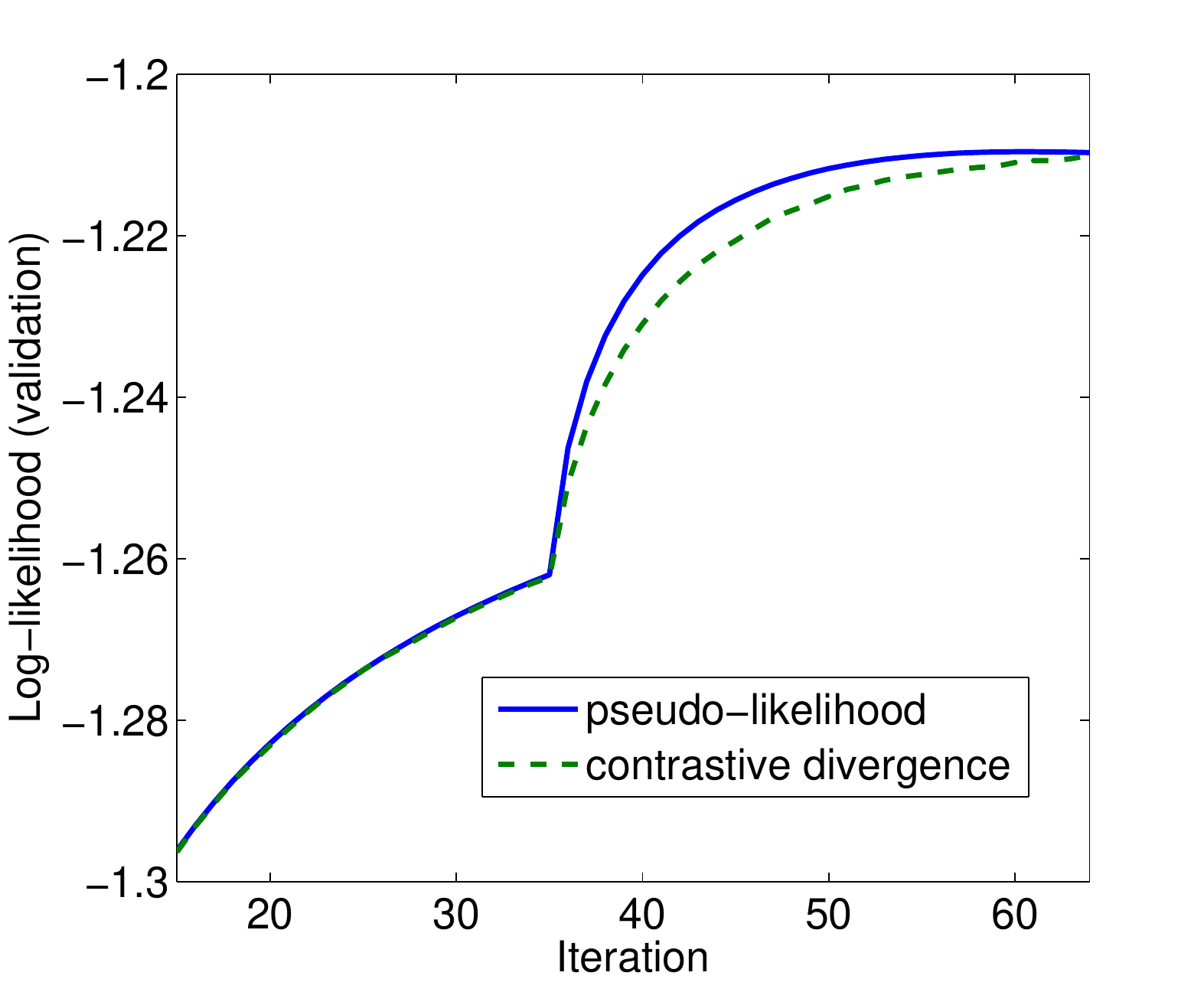} & \tabularnewline
$\lambda_{1}=10^{-3}$ & \tabularnewline
\end{tabular}
\par\end{centering}

\protect\caption{Learning curves for pseudo-likelihood (PL, Sec.~\ref{sub:Pseudo-likelihood})
versus contrastive divergence (CD, Sec\@.~\ref{sub:Contrastive-divergence}).
Data: MovieLens 1M; model: user-specific with \emph{smoothness parameterization}
(Sec.~\ref{sub:Smoothness-parameterization}).\label{fig:Learning-curve-PL-vs-CD}}
\end{figure}

Fig.~\ref{fig:Learning-curve} shows typical learning curves. There
is a gap between the pseudo-likelihoods on training and validation
data. The gap is widen as soon as pairwise parameters are introduced,
suggesting that there is a chance of overfitting. Thus the $\ell_{1}$-regularization
and validation for early stopping are essential. Increasing the penalty
factor from $\lambda_{1}=10^{-5}$(Fig.~\ref{fig:Learning-curve}-left)
to $\lambda_{1}=10^{-2}$ (Fig.~\ref{fig:Learning-curve}-right)
helps tremendously in combating against overfitting. Fig.~\ref{fig:Learning-curve-PL-vs-CD}
depicts comparison between pseudo-likelihood (Sec.~\ref{sub:Pseudo-likelihood})
and contrastive divergence (Sec.~\ref{sub:Contrastive-divergence}).
Overall these two learning methods behave similarly, with pseudo-likelihood
produces a slightly faster convergence. For the rest of the section,
we will report the results for pseudo-likelihood training only unless
specified otherwise.

\subsubsection{Sensitivity analysis}

\begin{figure}
\begin{centering}
\includegraphics[width=0.5\textwidth]{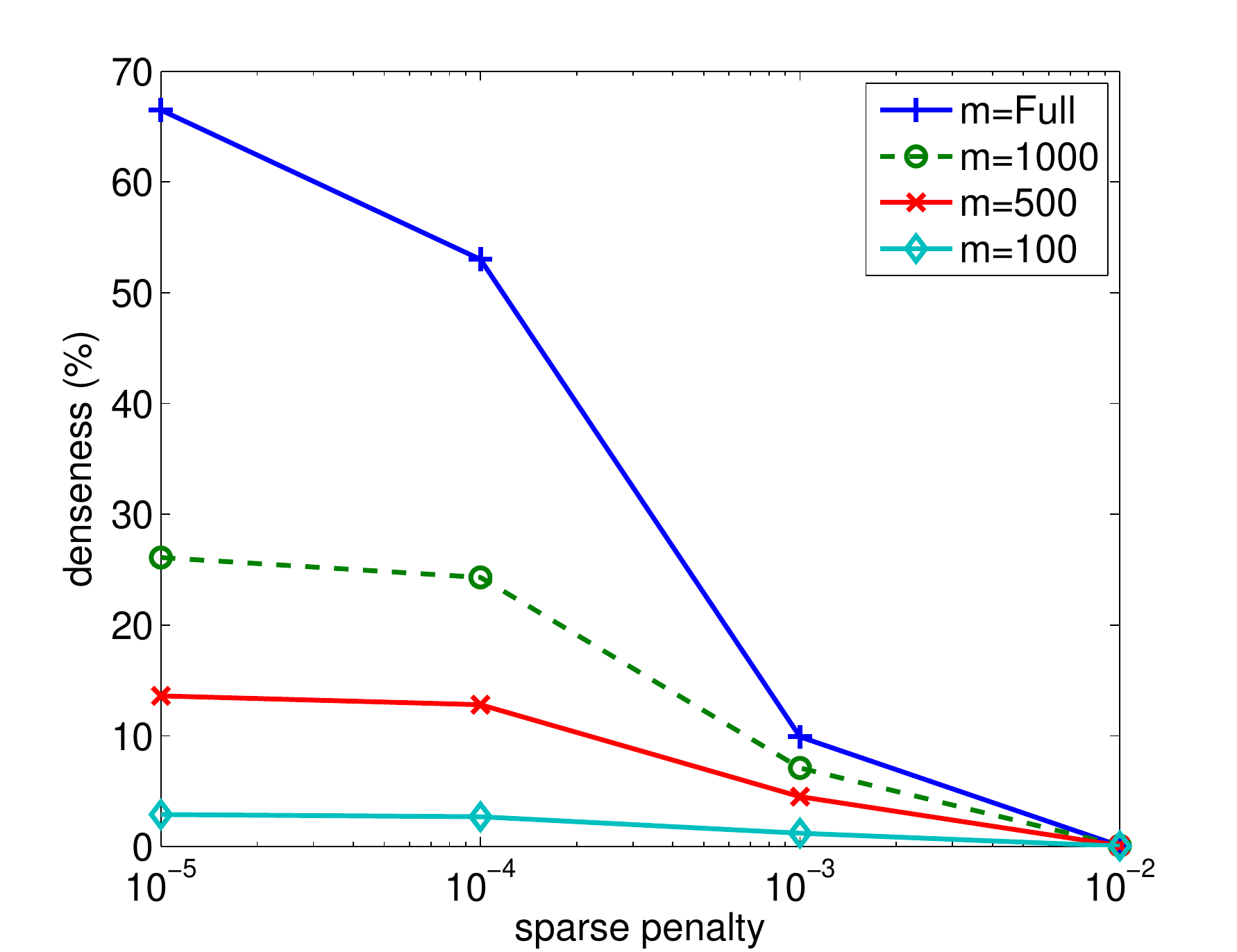}
\par\end{centering}

\protect\caption{Sparsity controlled by the $\ell_{1}$-penalty $\lambda_{1}$. Data:
MovieLens 1M; model: user-specific with \emph{smoothness parameterization
}(Sec.~\ref{sub:Smoothness-parameterization}) learnt from \emph{pseudo-likelihood}
(Sec.~\ref{sub:Pseudo-likelihood}).\label{fig:Sparsity-controlled-by}}
\end{figure}

\begin{figure}

\begin{centering}
\begin{tabular}{ccc}
\includegraphics[width=0.3\textwidth,height=0.3\textwidth]{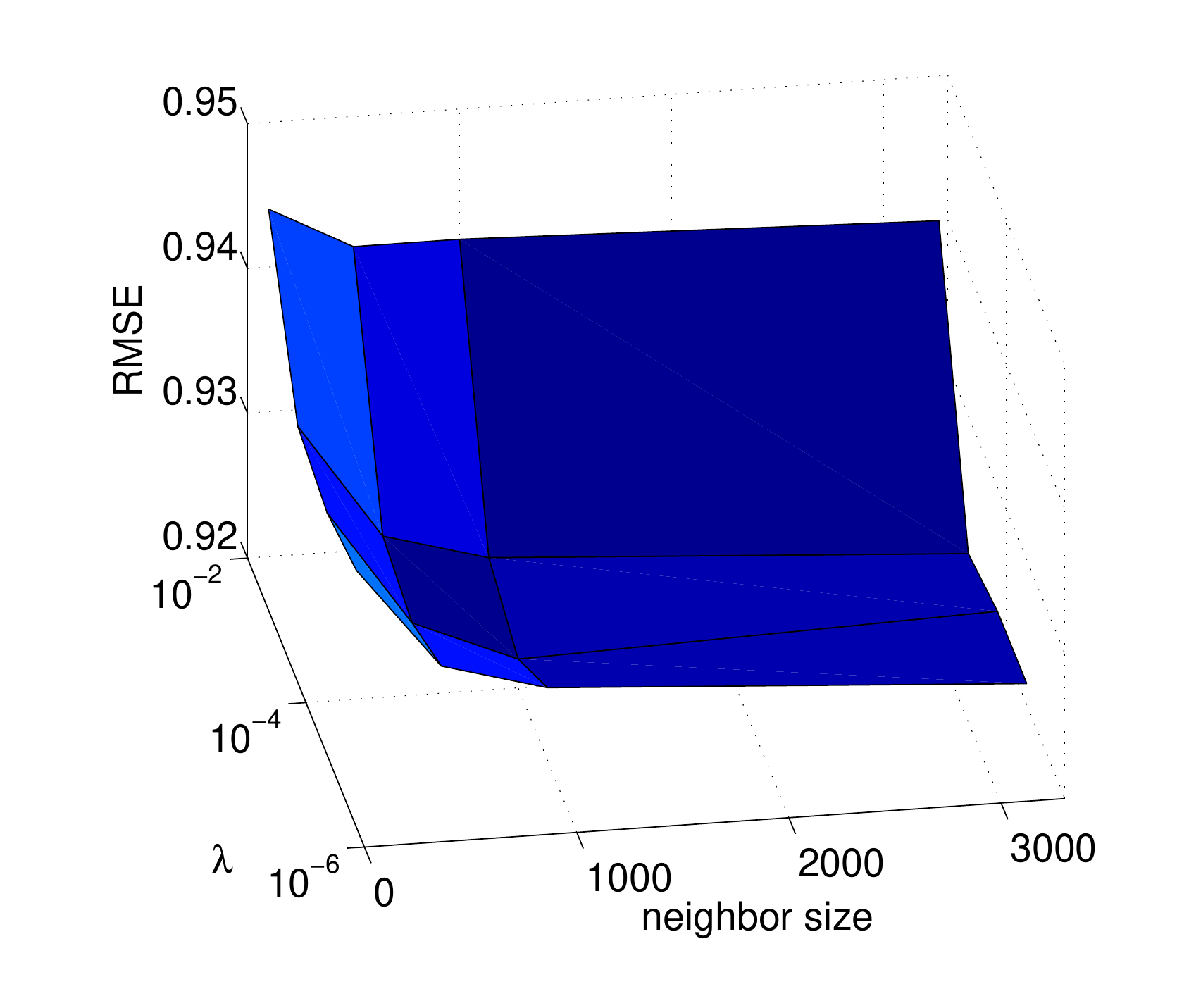} & \includegraphics[width=0.3\textwidth,height=0.3\textwidth]{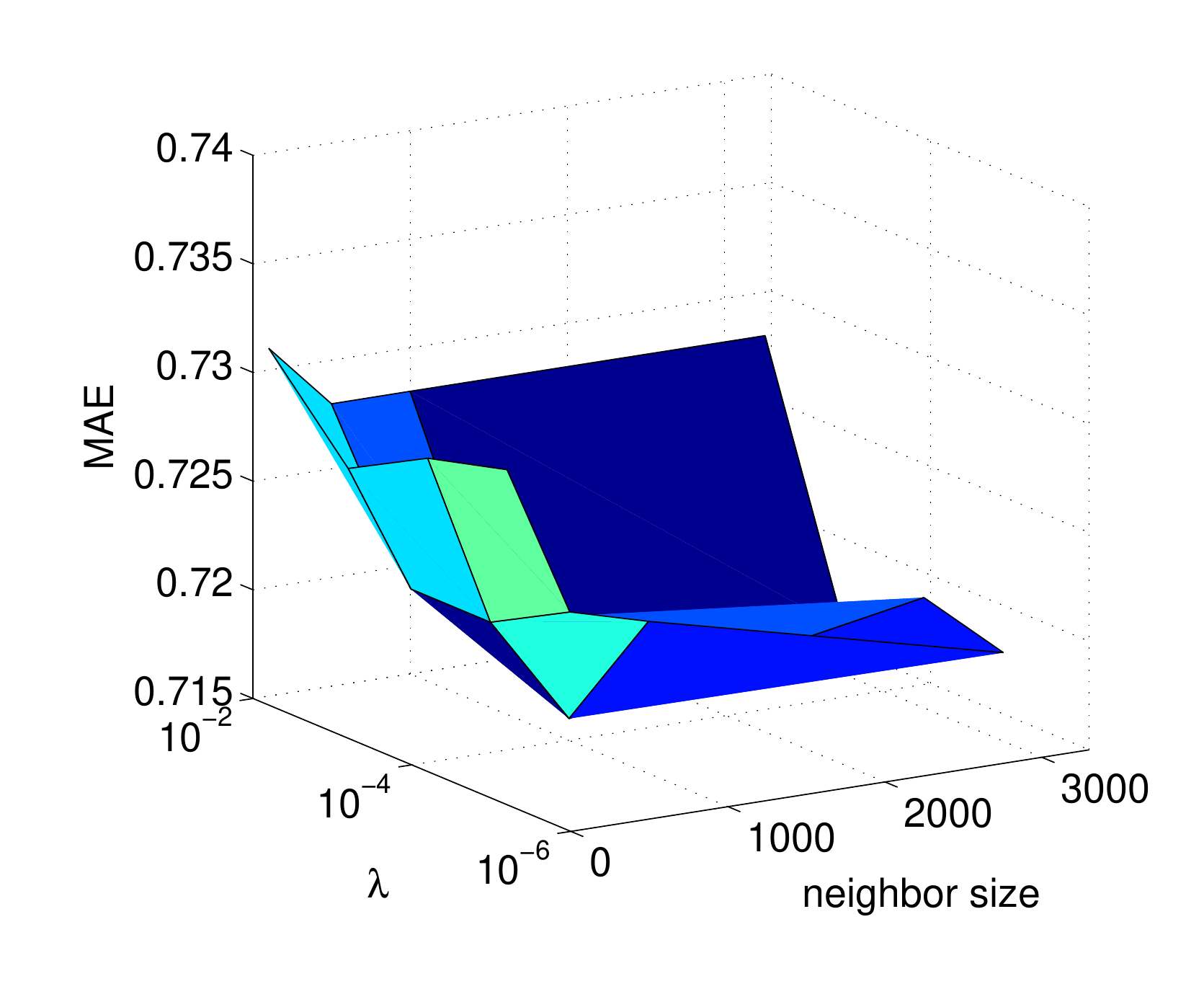} & \includegraphics[width=0.3\textwidth,height=0.3\textwidth]{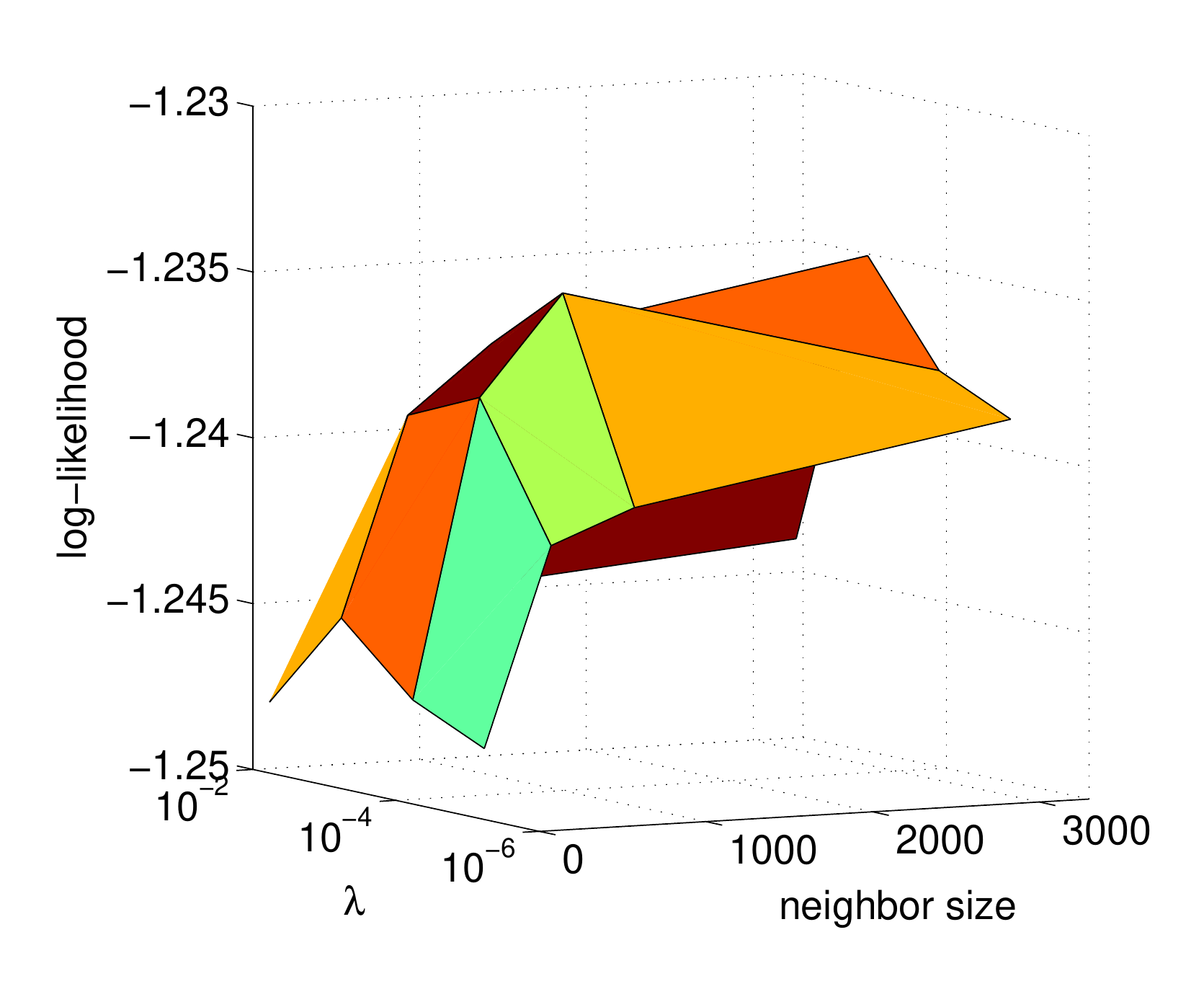}\tabularnewline
 &  & \tabularnewline
\end{tabular}
\par\end{centering}

\protect\caption{Performance sensitivity against hyperparameters: $m$ - max size of
neighborhood, and $\lambda_{1}$ - the $\ell_{1}$-penalty in Eq.~(\ref{eq:param-learn}).
Data: MovieLens 1M; model: user-specific with \emph{smoothness parameterization
}(Sec.~\ref{sub:Smoothness-parameterization}) learnt from \emph{pseudo-likelihood}
(Sec.~\ref{sub:Pseudo-likelihood}) (Best viewed in color).}

\end{figure}

\begin{figure}
\begin{centering}
\includegraphics[width=0.5\textwidth]{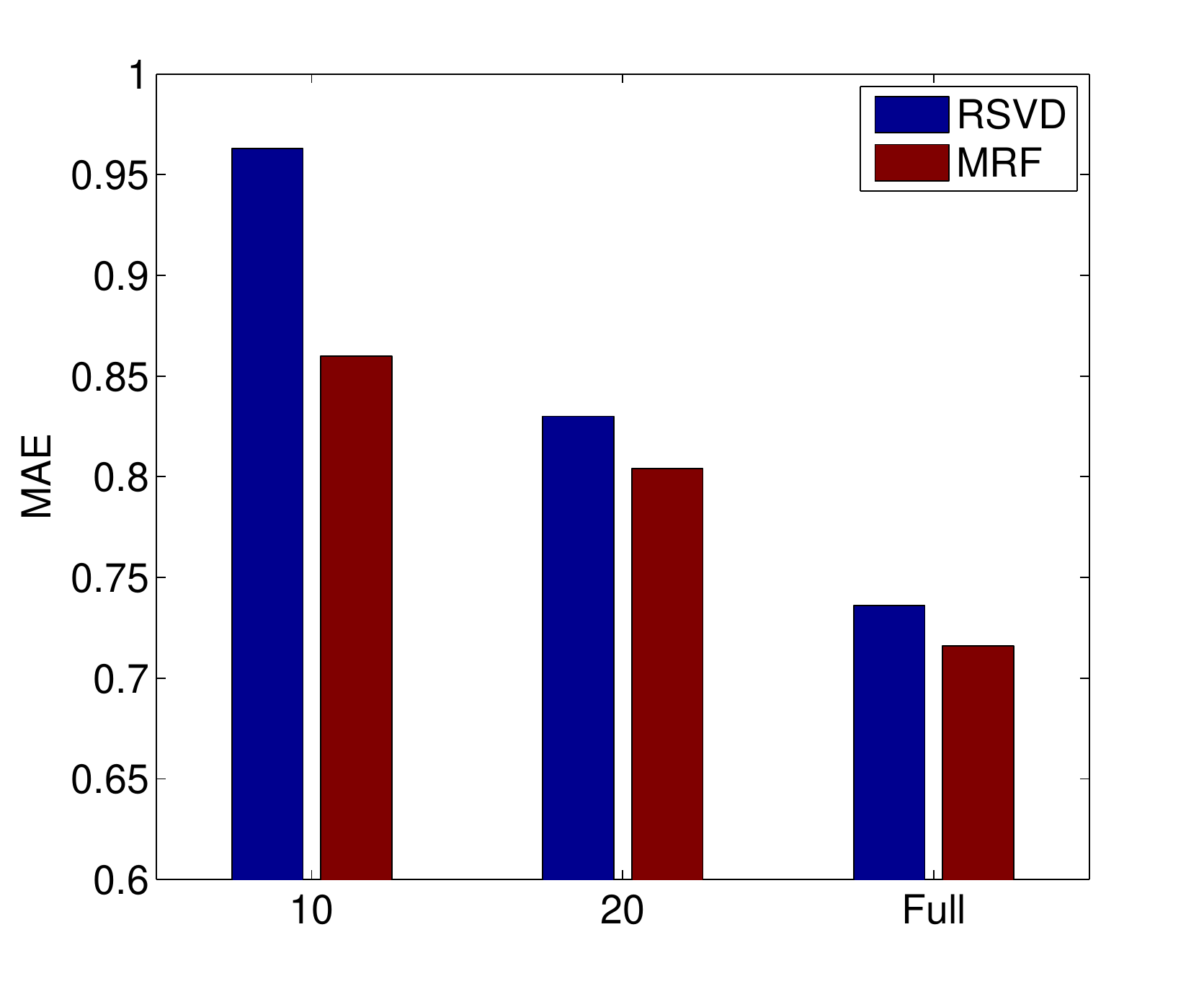}
\par\end{centering}

\protect\caption{Mean Absolute Error (MAE) as function of training size per user, $\lambda_{1}=10^{-3}$
(the smaller MAE the better). Data: MovieLens 1M; model: user-specific
with \emph{smoothness parameterization }(Sec.~\ref{sub:Smoothness-parameterization})
learnt from \emph{pseudo-likelihood} (Sec.~\ref{sub:Pseudo-likelihood})
(Best viewed in color).\label{fig:MAE-vs-train-size}}

\end{figure}

The sparsity of the graphs are measured as the ratio of number of
non-zeros edges and number of fully connected edges. Fig.~\ref{fig:Sparsity-controlled-by}
represents graph sparsity against the $\ell_{1}$-penalty $\lambda_{1}$
and the max neighborhood size $m$ (see Sec.~\ref{sub:Reducing-complexity}).
Larger penalty and smaller neighborhood size lead to more sparsity
(equivalently, less denseness). However, the two hyperparameters $\lambda_{1}$
and $m$ do affect the performance. For fully connected item graphs,
top performance is reached at $\lambda_{1}=10^{-3}$ on the MovieLens
1M data, achieving a sparsity of $9.9\%$. For smaller neighborhoods,
$\lambda_{1}=10^{-4}$ is the best setting. Overall, the performance
depends more on the sparsity penalty, and less on the neighborhood
size. It is desirable because it allows significant reduction of memory
footprint, which is proportional to $m$, with little loss of accuracy.

To verify whether the MRF can work with limited data, we randomly
pick $q$ ratings per user in the training data. Fig.~\ref{fig:MAE-vs-train-size}
depicts the behavior of the MRF with the smoothness parameterization
when $q=\left\{ 10,20,\mbox{full}\right\} $. The behavior is consistent
with the expectation that more data would lead to better performance.
Importantly, compared against the RSVD, it shows that the MRF is more
robust against small data.

\subsubsection{User and movie graphs}

\begin{figure}
\begin{centering}
\begin{tabular}{cc}
\includegraphics[width=0.45\textwidth,height=0.45\textwidth]{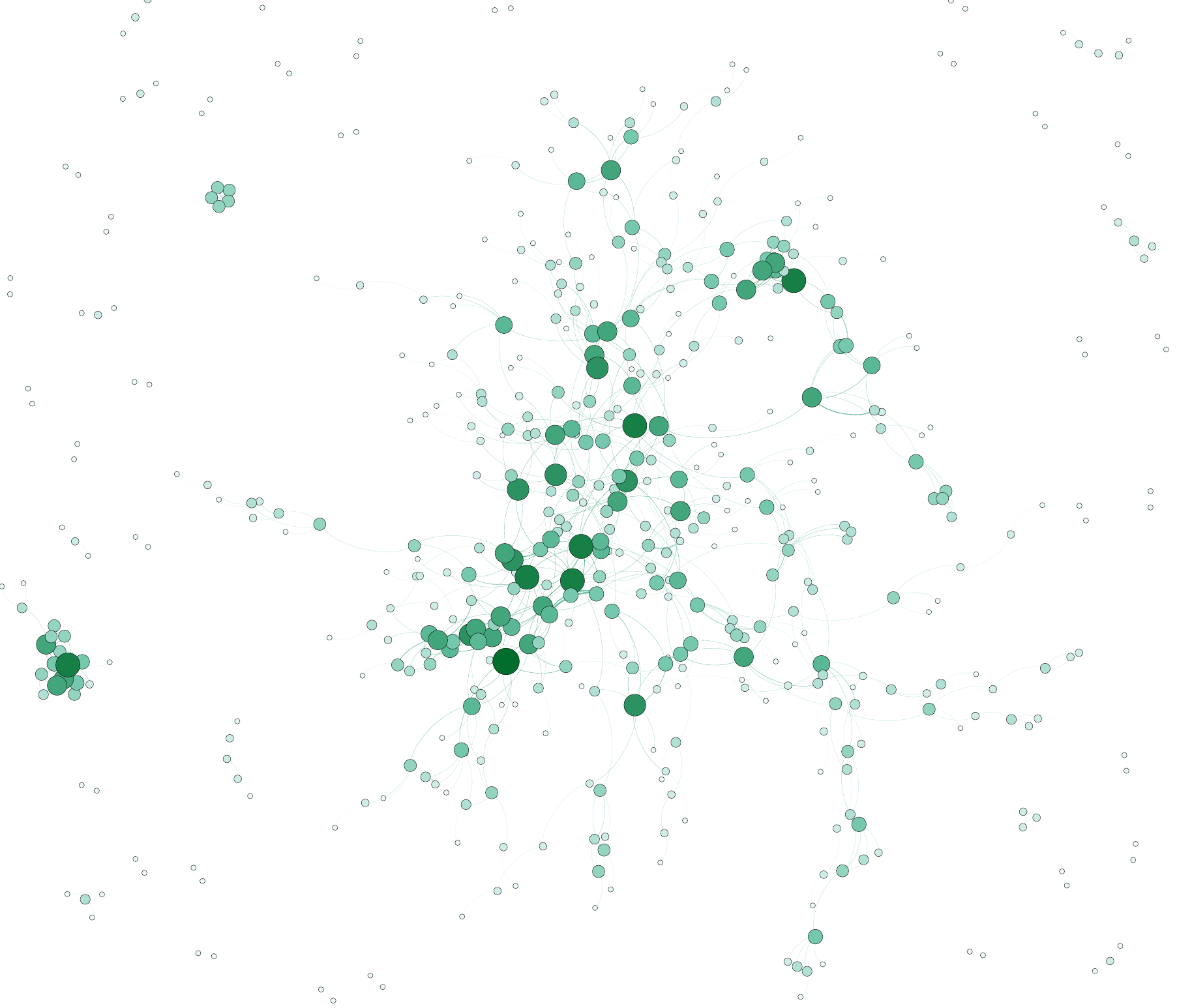}\quad{} & \quad \includegraphics[width=0.45\textwidth,height=0.45\textwidth]{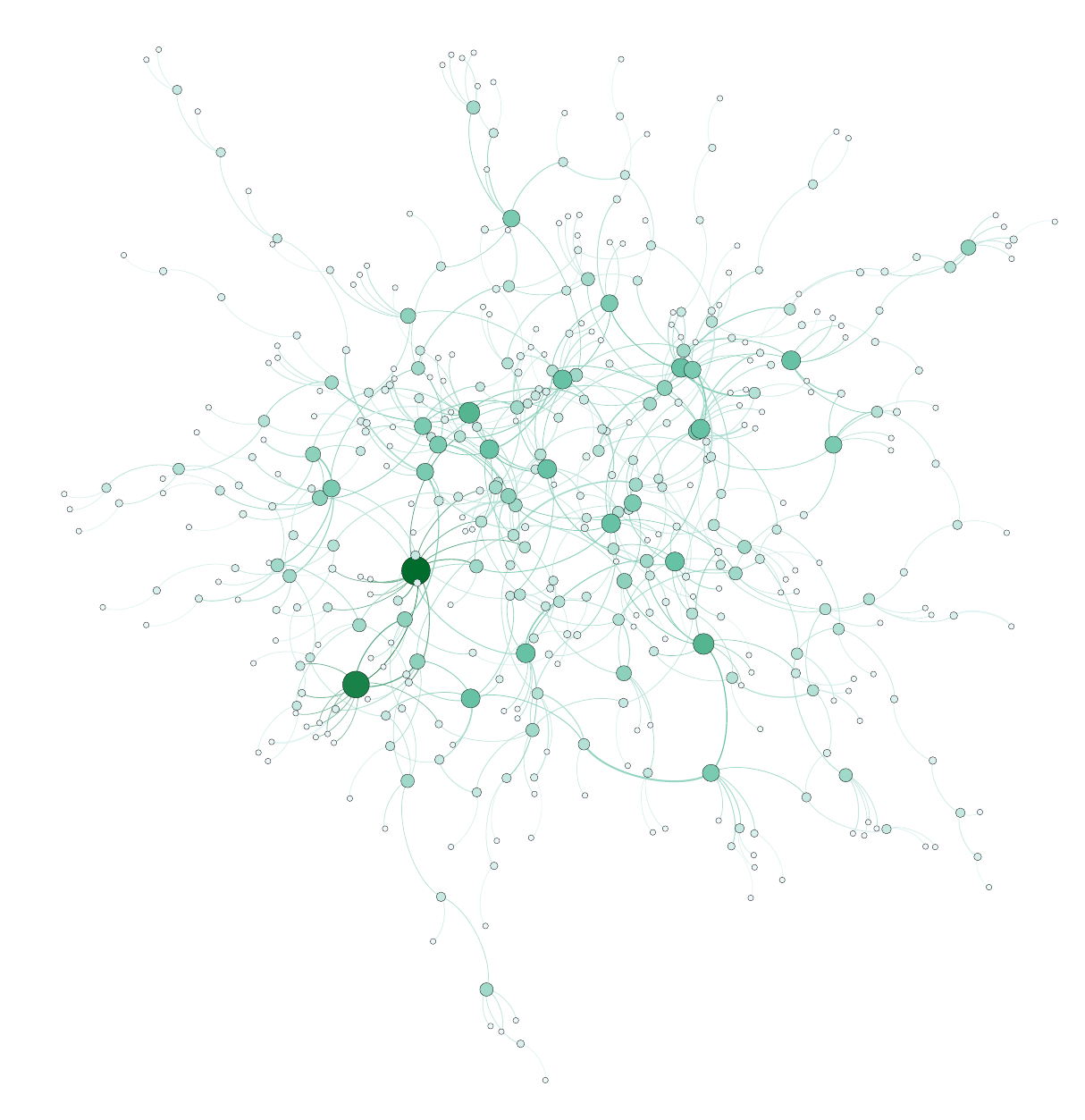}\tabularnewline
(a) Positive correlation & (a) Negative correlation\tabularnewline
\end{tabular}
\par\end{centering}

\protect\caption{Movie graphs ($\lambda_{1}=\lambda_{2}=10^{-3}$, smoothness parameterization,
contrastive divergence training). Node size represents connectivity
degree. To prevent clutters, only strong correlations are shown. \label{fig:Estimated-movie-graphs}}
\end{figure}

\begin{figure}
\begin{centering}
\begin{tabular}{cc}
\includegraphics[width=0.45\textwidth,height=0.45\textwidth]{user_correl_param_sparse2_MovieLens1M_1e-3_filter0\lyxdot 13max-positive}\quad{} & \quad \includegraphics[width=0.45\textwidth,height=0.45\textwidth]{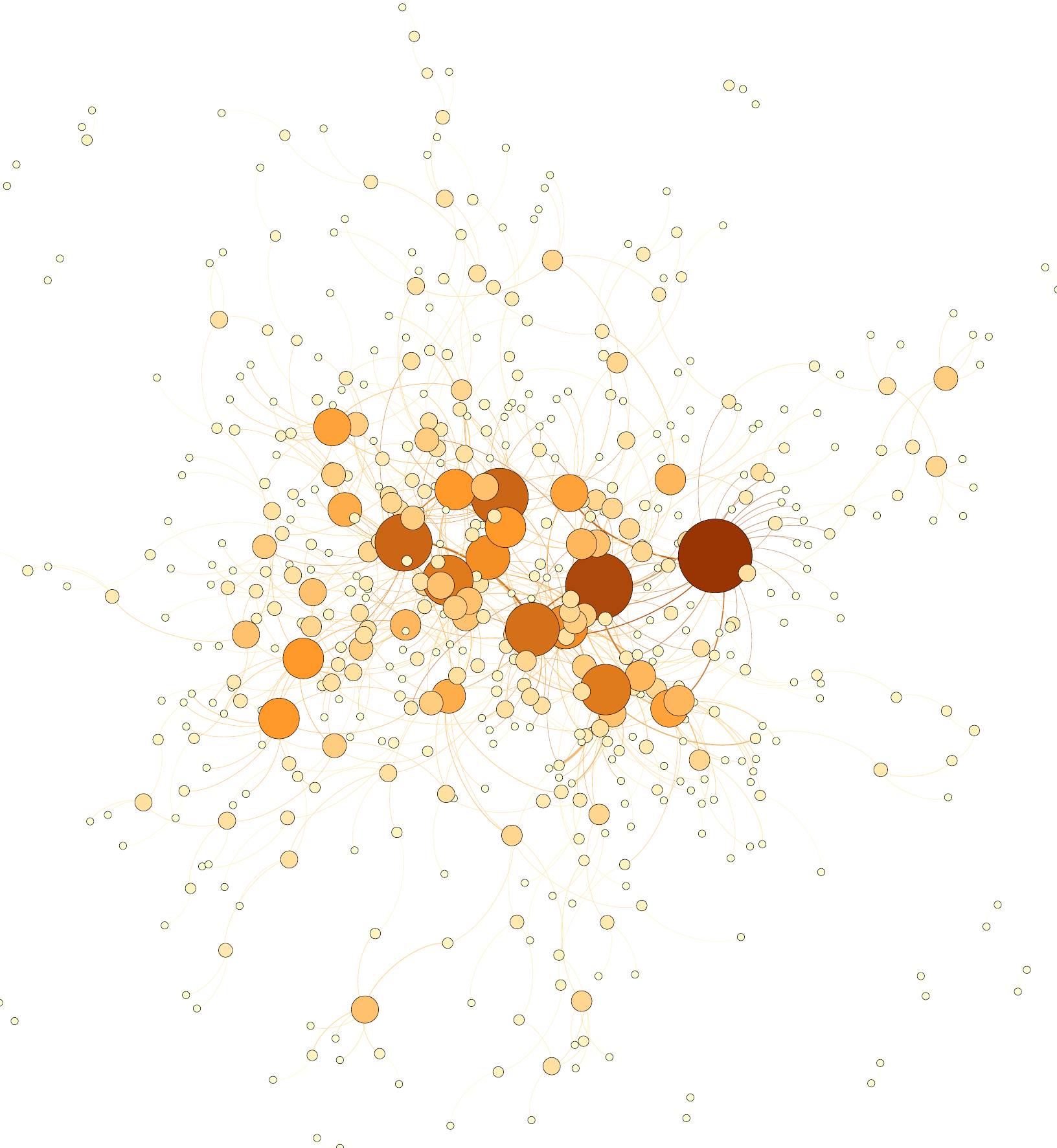}\tabularnewline
(a) Positive correlation & (a) Negative correlation\tabularnewline
\end{tabular}
\par\end{centering}

\protect\caption{User graphs on movie data ($\lambda_{1}=\lambda_{2}=10^{-3}$, smoothness
parameterization, contrastive divergence training). Node size represents
connectivity degree. To prevent clutters, only strong correlations
are shown. \label{fig:Estimated-user-graphs}}
\end{figure}

Our $\ell_{1}$-regularization framework in Eq.~(\ref{eq:param-learn-joint})
naturally discovers item correlation and user correlation graphs.
As previous demonstrated in Fig.~\ref{fig:Sparsity-controlled-by},
by varying the penalty hyperparameters $\lambda_{1}$ (to control
item-item graph sparsity) and $\lambda_{2}$ (to control user-user
graph sparsity), we obtain different connectivity patterns with varying
degree of sparsity. Figs.~\ref{fig:Estimated-movie-graphs}(a,b)
plot movie graphs estimated from the MovieLens 1M data. Positively
correlated movies are those liked (or disliked) in similar ways, while
a negative correlation means the two movies received diverse opinions.
The user graphs shown in Figs.~\ref{fig:Estimated-user-graphs} appear
to have a handful of users who have a high degree of connectivity,
either positively (agreement) or negatively (disagreement). This fact
could be exploited to locate influential users in social networks.

\subsubsection{Model performance}

Tab.~\ref{tab:Result-MovieLens} reports the results on the test
data for various model settings. The MRF's performance consistently
improves when user models and item models are joined. The Gaussian
parameterization (with careful normalization described in Sec.~\ref{sub:Gaussian-parameterization})
achieves the best fit in term of data log-likelihood (LL). All MRF
parameterizations excel on MAE and LL measures. The best performance
in RMSE and MAE are smoothness parameterization. 

\begin{table}
\begin{centering}
\begin{tabular}{|lccc|}
\hline 
\emph{Method} & \emph{RMSE} & \emph{MAE} & \emph{LL}\tabularnewline
\hline 
User-mean & 1.094 & 0.865 & -\tabularnewline
Item-mean & 1.006 & 0.806 & -\tabularnewline
Weighted-mean & 0.992 & 0.792 & -\tabularnewline
RSVD ($F=50$) & 0.932 & 0.736 & -1.353\tabularnewline
RSVD ($F=100$) & 0.921 & 0.729 & -1.311\tabularnewline
\hline 
MRF.\emph{user}.Gauss.PL & 0.930 & \textbf{0.728} & \textbf{-1.189}\tabularnewline
MRF.\emph{item}.Gauss.PL & 0.929 & \textbf{0.726} & \textbf{-1.188}\tabularnewline
MRF.\emph{joint}.Gauss.PL & 0.929 & \textbf{0.726} & \textbf{-1.188}\tabularnewline
\hline 
MRF.\emph{user}.linear-linear.PL & 0.940 & \textbf{0.724} & \textbf{-1.249}\tabularnewline
MRF.\emph{item}.linear-linear.PL & 0.942 & \textbf{0.720} & \textbf{-1.253}\tabularnewline
MRF.\emph{joint}.linear-linear.PL & 0.931 & \textbf{0.708} & \textbf{-1.243}\tabularnewline
\hline 
MRF.\emph{user}.smooth.PL & 0.922 & \textbf{0.716} & \textbf{-1.235}\tabularnewline
MRF.\emph{item}.smooth.PL & \textbf{0.920} & \textbf{0.712} & \textbf{-1.229}\tabularnewline
MRF.\emph{joint}.smooth.PL & \textbf{0.912} & \textbf{0.703} & \textbf{-1.218}\tabularnewline
\hline 
MRF.\emph{user}.smooth.CD & 0.922 & \textbf{0.716} & \textbf{-1.234}\tabularnewline
MRF.\emph{user}.smooth.CD & \textbf{0.914} & \textbf{0.704} & \textbf{-1.221}\tabularnewline
MRF.\emph{joint}.smooth.CD & \textbf{0.910} & \textbf{0.702} & \textbf{-1.216}\tabularnewline
\hline 
\end{tabular}
\par\end{centering}

\protect\caption{Rating prediction for MovieLens 1M, $m=M$ and $\lambda_{1}=\lambda_{2}=10^{-3}$.
$F$ is the number of hidden dimensions in the RSVD. Legend: RMSE
-- root mean squared error; MAE -- mean absolute error; LL -- data
log-likelihood; user -- user-specific model; item -- item-specific
model; joint -- joint model; PL -- pseudo-likelihood; and CD -- contrastive
divergence. Bolds indicate that the MRF-based results are better than
the baselines. \label{tab:Result-MovieLens}}

\end{table}

\subsection{Date matching}

For the date matching task, the dataset is Dating Agency\footnote{http://www.occamslab.com/petricek/data/}
with 17 million ratings in a $10$-point scale by $135$ thousand
users who rated nearly $169$ thousand profiles. To make the experiments
comparable with those in movie recommendation, we rescale the ratings
of the Dating Agency to the $5$-point scale. After processing, we
retain $101.3$ thousand users, $34.9$ thousand items and $12.5$
million ratings. The mean rating is $1.9$ (std: $1.5$) and the ratings
are quite uniformly distributed. The rating matrix is $0.4\%$ dense,
meaning that only $0.4\%$ profiles are rated on average. An average
user rates $123$ profiles but median is only $73$, indicating a
skew toward small rating history. An item is rated $357$ times on
average (median: $192$). As the number of users and profiles are
large, it is necessary to limit the neighborhood size to $m\ll M$
for manageable memory footprint. Other than that we use the same settings
as in the movie recommendation experiments. 

\begin{table}
\begin{centering}
\begin{tabular}{|l|lccc|}
\hline 
\emph{Method} & $m$ & \emph{RMSE} & \emph{MAE} & \emph{LL}\tabularnewline
\hline 
User-mean & - & 1.377 & 1.159 & -\tabularnewline
Item-mean & - & 0.932 & 0.654 & -\tabularnewline
Weighted-mean & - & 0.955 & 0.743 & -\tabularnewline
RSVD ($F=50$) & - & 0.886 & 0.616 & -1.084\tabularnewline
RSVD ($F=100$) & - & 0.895 & 0.624 & -1.096\tabularnewline
\hline 
MRF.\emph{user}.Gauss.PL & $1,000$ & \textbf{0.862} & \textbf{0.615} & \textbf{-1.150}\tabularnewline
MRF.\emph{user}.linear-linear.PL & $1,000$ & \textbf{0.847} & \textbf{0.517} & \textbf{-0.858}\tabularnewline
MRF.\emph{user}.smooth.PL & $1,000$ & \textbf{0.808} & \textbf{0.480} & \textbf{-0.821}\tabularnewline
MRF.\emph{user}.smooth.PL & $3,000$ & \textbf{0.793} & \textbf{0.467} & \textbf{-0.807}\tabularnewline
MRF.\emph{user}.smooth.PL & $5,000$ & \textbf{0.789} & \textbf{0.463} & \textbf{-0.803}\tabularnewline
\hline 
\end{tabular}
\par\end{centering}

\protect\caption{Rating prediction for Dating Agency data. Bolds indicate that the
MRF-based results are better than the baselines. Here:$\lambda_{1}=\lambda_{2}=10^{-5}$.\label{tab:Result-Dating}}
\end{table}

Tab.~\ref{tab:Result-Dating} reports the results. The simple item
mean performs surprisingly well (MAE: $0.654$) compared to the more
sophisticated method RSVD (MAE: $0.616$ with $K=50$). As the RMSE
essentially captures the variance of each method, the user variance
is much higher than item variance. It suggests that users are willing
to rate a diverse set of profiles. In addition, profiles receive a
high degree of agreement (e.g., with smaller RMSE and MAE, on average). 

As with the previous experiments on movie data, the smoothness parameterization
leads to the best performance. In particular, with $m=1,000$, the
user-specific model achieves a MAE of $0.480$, which is $22.1\%$
better than the best baseline (RSVD with $50$ hidden features). The
improvement increases to $24.8\%$ when $m$ is enlarged to $5,000$.
Note that this is still a small neighborhood, .e., $m=1,000$ accounts
for only $2.9\%$ of full item neighborhood.

Unlike the case of MovieLens 1M, the Gaussian parameterization does
not fit the data well. Interestingly, the data likelihood is similar
to that in the case of MovieLens 1M, regardless of the differences
between the two datasets. This could be due to unrealistic distribution
assumption about the unit variance and the full real-valued domain
for the normalized rating (the domain of the ratings in fact contains
only $5$ discrete points).

\section{Conclusion and future work \label{sec:Conclusion}}

This paper focuses on Markov random fields (MRF) as a principled method
for modeling a recommender system. We aimed to solve the open problem
of structure learning in the MRFs, which happen to be among the largest
networks ever studied, with millions of nodes and hundreds of millions
of edges. Our solution has two components. One is the log-linear parameterization
schemes and the other is a sparsity-inducing framework through $\ell_{1}$-norm
regularization. Unlike existing work where model structure must be
specified by hand, our framework jointly discovers item-item and user-user
networks from data in a principled manner. The density of these networks
can be easily controlled by a hyper-parameter. We evaluated the proposal
through extensive experiments on two large-scale datasets -- the MovieLens
1M with 1 million ratings, the Dating Agency with 17 million ratings.

\subsection{Findings}

The experiments lead to the following findings:
\begin{itemize}
\item Compared to state-of-the-art collaborative filtering algorithms, our
sparse MRFs have higher performance, and is more robust against small
training data.
\item There exist optimal sparsity factors $\lambda_{1}$ and $\lambda_{2}$
(see Eq.~(\ref{eq:param-learn-joint})), with respect to prediction
accuracy.
\item The complexity of the learning algorithm can be significantly reduced
by several orders of magnitude through selecting a small neighborhood
size with little loss of accuracy.
\item Generally, the smoothness parameterization (Sec.~\ref{sub:Smoothness-parameterization})
does best in RMSE and MAE.
\item For the MovieLens 1M data, the user graphs have the ``hubness''
characteristic, where there exist several users with high degree of
connectivity (e.g., see \cite{tomavsev2011role}). They are likely
to be the influencers in this social network.
\item Finally, a compact and powerful MRF can be estimated efficiently for
recommender systems which may involve hundreds of millions of parameters.
\end{itemize}

\subsection{Limitations and future work}

We observe several limitations which open rooms for future work:
\begin{itemize}
\item The MRFs, while powerful and accurate, are expensive to train, compared
to the latent aspects approach RSVD and RBM. We have introduced a
way to reduce the complexity significantly by using only popular neighbors
without hurting the performance. Further, MRFs can be combined with
RSVD in our recent work in \cite{liu2014ordinal}, and combined with
RBM in \cite{Truyen:2009a} but the sparse MRFs have not been investigated.
\item This paper, like the majority of collaborative filtering literature,
assumes that recommendations are first based on rating prediction.
While it is reasonable to assume that we should recommend items with
potentially high personal rating, it ignores other dimensions such
as novelty and diversity. One solution is to use entropy as a measure
of novelty:
\[
H(u,i)=-\sum_{r_{ui}}P\left(r_{ui}\mid\rb\right)\log P\left(r_{ui}\mid\rb\right)
\]

\item Since the attention of the user is limited, it is better to suggest
just a few items at a time. As such, item ranking may be more appropriate
than rate prediction. Motivated by the expected rating in Eq.~(\ref{eq:MRF-rate-expect}),
we propose to use expected energy decrease as ranking criterion
\begin{equation}
s_{j}=\sum_{r_{uj}}P\left(r_{uj}\mid\rb_{N(j)}\right)\left[-E\left(r_{uj},\rb_{N(j)}\right)\right]\label{eq:energy-change}
\end{equation}
The motivation behind this criterion is the observation that when
a new item is added to an user's list, the energy of the system decreases
if the item is compatible with the user. Thus, the lower the energy,
the more preferable item. The same argument leads to another criterion
-- the change in free-energy:
\begin{equation}
s_{j}'=\sum_{r_{uj}}\exp\left(-E\left(r_{uj},\rb_{N(j)}\right)\right)\label{eq:free-energy-change}
\end{equation}
In \cite{li2015Preference}, a MRF based solution has been introduced
for ranking, but without sparse MRFs.
\item An undesirable effect of the hubness property found in practice is
that for some hub users and items, the conditional distribution $P\left(r_{ui}\mid\ratemat_{\neg ui}\right)$
could be peaked due to many contributions from neighbors. Our bias
handling in Eqs.~(\ref{eq:pairwise-poten-multinomial-user-user},\ref{eq:pairwise-poten-multinomial-item-item},\ref{eq:pairwise-poten-gauss-user-user},\ref{eq:pairwise-poten-gauss-item-item},\ref{eq:pairwise-poten-ord-user-user},\ref{eq:pairwise-poten-ord-item-item})
has partly mitigate the problem. For sparsely connected users and
items, we can assume that the unseen ratings are the mean rating,
thus the pairwise potentials that link with those unseen ratings are
close to unity, i.e., $\psi_{ij},\varphi_{uv}\approx1$. Alternatively,
we could normalize the energy function against the size of the neighborhood
size. However, we found that this technique has little effect on the
final prediction performance.
\item Finally, it might be useful incorporate social graphs, or item graphs
learnt from external sources into our framework.\end{itemize}

\section*{References}

\bibliographystyle{plain}

\begin{thebibliography}{10}

\bibitem{agresti1990cda}
A.~Agresti.
\newblock {\em {Categorical data analysis}}.
\newblock Wiley-Interscience, 1990.

\bibitem{ali2004tivo}
K.~Ali and W.~Van~Stam.
\newblock {TiVo: making show recommendations using a distributed collaborative
  filtering architecture}.
\newblock In {\em Proceedings of the tenth ACM SIGKDD international conference
  on Knowledge discovery and data mining}, pages 394--401. ACM, 2004.

\bibitem{Besag-74}
Julian Besag.
\newblock Spatial interaction and the statistical analysis of lattice systems
  (with discussions).
\newblock {\em Journal of the Royal Statistical Society Series B}, 36:192--236,
  1974.

\bibitem{das2007gnp}
A.S. Das, M.~Datar, A.~Garg, and S.~Rajaram.
\newblock {Google news personalization: scalable online collaborative
  filtering}.
\newblock In {\em Proceedings of the 16th international conference on World
  Wide Web (WWW)}, pages 271--280. ACM Press New York, NY, USA, 2007.

\bibitem{defazio2012graphical}
Aaron Defazio and Tib{\'e}rio~S Caetano.
\newblock A graphical model formulation of collaborative filtering
  neighbourhood methods with fast maximum entropy training.
\newblock In {\em Proceedings of the 29th International Conference on Machine
  Learning (ICML-12)}, pages 265--272, 2012.

\bibitem{farzan2006social}
Rosta Farzan and Peter Brusilovsky.
\newblock Social navigation support in a course recommendation system.
\newblock In {\em Adaptive hypermedia and adaptive web-based systems}, pages
  91--100. Springer, 2006.

\bibitem{gunawardana2008tied}
Asela Gunawardana and Christopher Meek.
\newblock {Tied Boltzmann machines for cold start recommendations}.
\newblock In {\em Proceedings of the 2008 ACM conference on Recommender
  systems}, pages 19--26. ACM, 2008.

\bibitem{Hammersley-Clifford71}
J.M. Hammersley and P.~Clifford.
\newblock Markov fields on finite graphs and lattices.
\newblock Unpublished manuscript, 1971.

\bibitem{heckerman2001dni}
D.~Heckerman, D.M. Chickering, C.~Meek, R.~Rounthwaite, and C.~Kadie.
\newblock {Dependency networks for inference, collaborative filtering, and data
  visualization}.
\newblock {\em The Journal of Machine Learning Research}, 1:49--75, 2001.

\bibitem{Hinton02}
G.E. Hinton.
\newblock Training products of experts by minimizing contrastive divergence.
\newblock {\em Neural Computation}, 14:1771--1800, 2002.

\bibitem{hofmann2004lsm}
T.~Hofmann.
\newblock {Latent semantic models for collaborative filtering}.
\newblock {\em ACM Transactions on Information Systems (TOIS)}, 22(1):89--115,
  2004.

\bibitem{koren2010factor}
Y.~Koren.
\newblock Factor in the neighbors: Scalable and accurate collaborative
  filtering.
\newblock {\em ACM Transactions on Knowledge Discovery from Data (TKDD)},
  4(1):1, 2010.

\bibitem{linden2003amazon}
Greg Linden, Brent Smith, and Jeremy York.
\newblock {Amazon.com recommendations: Item-to-item collaborative filtering}.
\newblock {\em IEEE Internet Computing}, 7(1):76--80, 2003.

\bibitem{li2015Preference}
Shaowu Liu, Gang Li, Truyen Tran, and J~Yuan.
\newblock {Preference Relation-based Markov Random Fields}.
\newblock In {\em Proc. of 7th Asian Conference on Machine Learning (ACML)},
  Hongkong, November 2015.

\bibitem{liu2014ordinal}
Shaowu Liu, Truyen Tran, Gang Li, and J~Yuan.
\newblock Ordinal random fields for recommender systems.
\newblock In {\em Proc. of 6th Asian Conference on Machine Learning (ACML)},
  Nha Trang, Vietnam, November 2014.

\bibitem{lu2015recommender}
Jie Lu, Dianshuang Wu, Mingsong Mao, Wei Wang, and Guangquan Zhang.
\newblock {Recommender system application developments: A survey}.
\newblock {\em Decision Support Systems}, 74:12--32, 2015.

\bibitem{marlin2004mur}
B.~Marlin.
\newblock {Modeling user rating profiles for collaborative filtering}.
\newblock In {\em Advances in Neural Information Processing Systems},
  volume~16, pages 627--634. MIT Press, Cambridge, MA, 2004.

\bibitem{martinez2015model}
C~Martinez-Cruz, C~Porcel, J~Bernab{\'e}-Moreno, and E~Herrera-Viedma.
\newblock A model to represent users trust in recommender systems using
  ontologies and fuzzy linguistic modeling.
\newblock {\em Information Sciences}, 311:102--118, 2015.

\bibitem{resnick94grouplens}
P.~Resnick, N.~Iacovou, M.~Suchak, P.~Bergstorm, and J.~Riedl.
\newblock {GroupLens}: An open architecture for collaborative filtering of
  netnews.
\newblock In {\em Proceedings of {ACM} Conference on Computer Supported
  Cooperative Work}, pages 175--186, Chapel Hill, North Carolina, 1994. ACM.

\bibitem{salakhutdinov2008probabilistic}
R.~Salakhutdinov and A.~Mnih.
\newblock {Probabilistic matrix factorization}.
\newblock {\em Advances in neural information processing systems},
  20:1257--1264, 2008.

\bibitem{Salakhutdinov-et-alICML07}
R.~Salakhutdinov, A.~Mnih, and G.~Hinton.
\newblock {Restricted Boltzmann machines for collaborative filtering}.
\newblock In {\em Proceedings of the 24th ICML}, pages 791--798, 2007.

\bibitem{sarwar2001ibc}
B.~Sarwar, G.~Karypis, J.~Konstan, and J.~Reidl.
\newblock {Item-based collaborative filtering recommendation algorithms}.
\newblock In {\em Proceedings of the 10th international conference on World
  Wide Web}, pages 285--295. ACM Press New York, NY, USA, 2001.

\bibitem{tejeda2015refore}
A~Tejeda-Lorente, C~Porcel, J~Bernab{\'e}-Moreno, and E~Herrera-Viedma.
\newblock {REFORE: A recommender system for researchers based on
  bibliometrics}.
\newblock {\em Applied Soft Computing}, 30:778--791, 2015.

\bibitem{tomavsev2011role}
Nenad Toma{\v{s}}ev, Milo{\v{s}} Radovanovi{\'c}, Dunja Mladeni{\'c}, and
  Mirjana Ivanovi{\'c}.
\newblock The role of hubness in clustering high-dimensional data.
\newblock In {\em Advances in Knowledge Discovery and Data Mining}, pages
  183--195. Springer, 2011.

\bibitem{Truyen:2007}
T.T. Truyen, D.Q. Phung, and S.~Venkatesh.
\newblock {Preference networks: Probabilistic models for recommendation
  systems}.
\newblock In P.~Christen, P.J. Kennedy, J.~Li, I.~Kolyshkina, and G.J.
  Williams, editors, {\em The 6th Australasian Data Mining Conference (AusDM)},
  volume~70 of {\em CRPIT}, pages 195--202, Gold Coast, Australia, Dec 2007.
  ACS.

\bibitem{Truyen:2009a}
T.T. Truyen, D.Q. Phung, and S.~Venkatesh.
\newblock {Ordinal Boltzmann machines for collaborative filtering}.
\newblock In {\em Twenty-Fifth Conference on Uncertainty in Artificial
  Intelligence (UAI)}, Montreal, Canada, June 2009.

\bibitem{wang2006unifying}
Jun Wang, Arjen~P De~Vries, and Marcel~JT Reinders.
\newblock Unifying user-based and item-based collaborative filtering approaches
  by similarity fusion.
\newblock In {\em Proceedings of the 29th annual international ACM SIGIR
  conference on Research and development in information retrieval}, pages
  501--508. ACM, 2006.

\bibitem{zou2013iterative}
Jun Zou, Arash Einolghozati, Erman Ayday, and Faramarz Fekri.
\newblock Iterative similarity inference via message passing in factor graphs
  for collaborative filtering.
\newblock In {\em Information Theory Workshop (ITW), 2013 IEEE}, pages 1--5.
  IEEE, 2013.

\end{thebibliography}

\end{document}